\documentclass{article}
\usepackage{spconf,amsmath,graphicx}
\usepackage{cite}
\usepackage{amsmath,amssymb,amsfonts}
\usepackage{algorithmic}
\usepackage{graphicx}
\usepackage{textcomp}
\usepackage{xcolor}
\usepackage{booktabs}
\usepackage{makecell}
\usepackage{soul}
\usepackage{subcaption}
\usepackage{hyperref}

\newcommand{\red}[1]{\textcolor{red}{#1}}

\title{SLACK: Attacking LiDAR-based SLAM with Adversarial Point Injections}
\name{Prashant Kumar$^{*1}$, Dheeraj Vattikonda$^{2}$, Kshitij Madhav Bhat$^{3}$,
Kunal Dargan$^{1}$,
Prem Kalra$^{1}$\thanks{Email: prashantk.nan@gmail.com; \href{https://drive.google.com/drive/folders/1BtrTwUbgFKYkIQ6UGitQAuClHH-pHjBH?usp=sharing}{Supplementary}}}
\address{IIT Delhi$^{1}$, McGill University$^{2}$, IIT Indore$^{3}$}

\newcommand{\atlas}{{\textsc{AtlasNet}}}
\newcommand{\Ach}{{\textsc{Achlioptas et al.}}}

\newcommand{\CacciaAE}{{\textsc{Caccia-AE}}}
\newcommand{\SAT}{{\textsc{SLACK}}}
\newcommand{\CP}{{\textsc{CP3}}}
\newcommand{\SATUDA}{{\textsc{SLACK-MMD}}}
\newcommand{\CacciaVAE}{{\textsc{Caccia-VAE}}}
\newcommand{\CacciaGAN}{{\textsc{Caccia-GAN}}}

\newcommand{\DSLR}{\textsc{DSLR}}
\newcommand{\RR}{\textsc{RR}}
\newcommand{\RN}{\textsc{RN}}

\begin{document}
%
\maketitle
\begin{abstract}

    The widespread adoption of learning-based methods  
for the LiDAR makes autonomous vehicles vulnerable to adversarial
attacks through adversarial \textit{point injections (PiJ)}. It poses serious security
challenges for navigation and map generation. Despite its
critical nature, no major work exists that studies learning-based attacks on LiDAR-based SLAM. Our work proposes SLACK,
an end-to-end deep generative adversarial model to attack LiDAR scans
with several point injections without deteriorating LiDAR
quality. To facilitate SLACK, we design a novel yet simple autoencoder that augments contrastive learning with segmentation-based attention for precise reconstructions. SLACK demonstrates superior performance on
the task of \textit{point injections (PiJ)} compared to the best baselines on KITTI and
CARLA-64 dataset while maintaining accurate scan quality. We qualitatively and quantitatively demonstrate PiJ
attacks using a fraction of LiDAR points. It severely degrades navigation and map quality without deteriorating the LiDAR scan quality.
\end{abstract}
\section{Introduction}

The integration of Autonomous Vehicles (AV) into our transportation system holds immense promise for increased safety and efficiency. However, technological leap requires robust security measures to address potential vulnerabilities. There is a growing concern about the interaction of intelligent systems and the web through over-the-air (OTA) updates; once an adversary gains access to the LiDAR preprocessing module, it can exploit point injections (PiJ) attacks\cite{liu2020computing,bgf:chattopadhyay2020autonomous}.
These manipulations, while minimal, can significantly disrupt the car's navigation system.

Despite the potential for disruption,
the challenges, methods, and impact of adversarial attacks on LiDAR-based SLAM have not been extensively investigated. Further research is crucial to develop robust defences against these emerging threats. Demonstrating and evaluating the impact of such attacks on LiDAR point clouds is extremely important to draw the attention of the Autonomous Vehicle (AV) community to these scenarios.

Cao et. al. \cite{cao2019adversarial} used specialized hardware for PiJ  
to manipulate individual laser beams and refract them to a wider angle.
However, hardware limitations restrict the number of fake points injected in a LiDAR scan \cite{sun2020towards}. In contrast to many existing adversarial attacks that target individual navigation modules like object detection and segmentation, our research addresses a more fundamental challenge i.e. compromising LiDAR-based SLAM navigation through the deliberate spurious point injection (PiJ)  into the LiDAR system. Rather than focusing solely on local structures and regions, we concentrate on subtly augmenting or tweaking the LiDAR scan with minimal point injections designed to destabilize navigation while ensuring the integrity of the LiDAR data. These injections target strategic regions, such as static structures crucial for SLAM.
Attacks on navigation systems may include passive attacks that affect a submodule assisting navigation - e.g. object detection \cite{cao2019adversarial,cho2023adopt}. On the contrary, ours is a white box threat model - we attack the navigation system by attack the SLAM system with erroneous LiDAR scans. These result in sub-optimal trajectory estimates. SLAM algorithms are only as good as the precision of the LiDAR scans fed to them. Attacking a LIDAR scan with adversarial noise hampers the trajectory estimates of the SLAM algorithm during navigation. 

By focusing on white-box PiJ attacks through network vulnerabilities,
 this research emphasizes a more realistic and concerning threat to the security of autonomous vehicles. It highlights the importance of securing data transmission and implementing robust detection systems to prevent Point injection (PiJ) manipulation.


\begin{figure}[htbp]

\begin{tabular}{c c}
    \includegraphics[width=0.47\linewidth, height=0.2\textwidth]{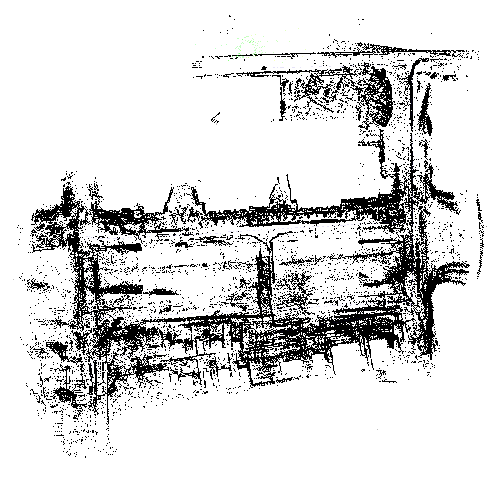} & \includegraphics[width=0.5\linewidth]{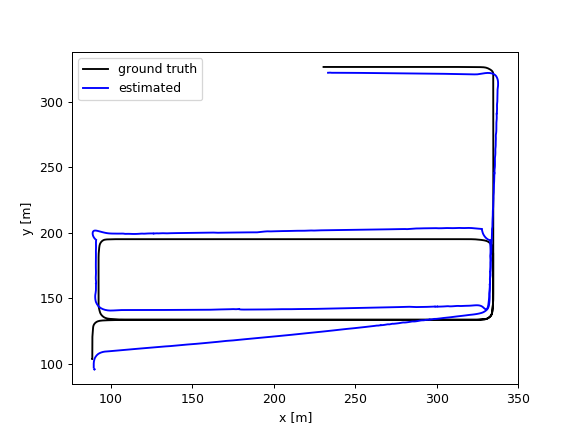}
\end{tabular}
\caption{\textbf{Left:} SLAM results before attack using CARLA sequences - the map is precise. Navigation trajectory is accurate. \textbf{Right:} After adversarial attack with 0.8\% points - map quality is degraded. High navigation error in estimated trajectory.}
\label{fig:slam-first}
\end{figure}

We approach this problem from a learning perspective. Our goal is to augment LiDAR with point injections (PiJ) without deteriorating the LiDAR scan quality. We ensure that the attack is difficult to detect but is strong enough to deteriorate navigation and map quality. To achieve this, we develop a novel, yet simple autoencoder backbone that uses segmentation-based attention coupled with contrastive learning using carefully chosen hard negatives. It enables our model to maintain LiDAR quality while camouflaging dynamic injections in it. This autoencoder combined with a pretext task discriminator forms our adversarial model. It injects a LiDAR scan with multiple dynamic points while maintaining the overall quality of a LiDAR scan. The injection is sufficient to attack LiDAR scans (PiJ) and severely affects navigation accuracy and map quality (Figure \ref{fig:slam-first}).
\\
We summarize the contributions of our paper as follows:\\

     $\bullet$ We demonstrate injection attacks using learning strategies in LiDAR point clouds. This is of critical importance for security assessment, navigation performance and map generation of AVs'.\\
    $\bullet$ We design a novel autoencoder backbone, $AE_{mask}$ that is a part of our adversarial module. $AE_{mask}$  uses binary segmentation-assisted attention coupled with contrastive learning using hard negatives. It achieves precise LiDAR reconstruction and preserves LiDAR quality. $AE_{mask}$ can be independently used as a backbone for numerous different generative modelling tasks.\\
     $\bullet$ To simulate PiJ attacks we develop \SAT - it combines $AE_{mask}$ with a novel pretext-task discriminator $PD$, in an adversarial fashion. It injects LiDAR scans with point injections that are sufficient enough to degrade SLAM. The attacked LiDAR is hard to differentiate from the original LiDAR.\\
     $\bullet$ Our adversarial model requires paired correspondence of dynamic-static scans. This may not be available for certain datasets. To overcome this, we propose \SATUDA{} that utilizes Unsupervised Domain Adaptation to demonstrate  PiJ attacks on these datasets. \\
     $\bullet$ We demonstrate PiJ attacks on the real-world KITTI and the simulated CARLA dataset sequences. The navigation introduced due to the attakcs is high enough to destabilize navigation. We also demonstrate severe deterioration in the quality of the generated map.

\section{Problem Formulation}
Our objective is to inject a LiDAR scan with few dynamic points injections sufficient enough to destabilize SLAM-based navigation without deteriorating scan quality. For this purpose, we use the simulated corresponding dynamic-static paired dataset, CARLA-64 and real-world KITTI and ARD-16 datasets. A corresponding LiDAR scan pair refers to a dynamic-static pair such that both scans are captured in the same location but the dynamic scan has dynamic objects while the static scan is devoid of dynamic objects.

Consider dynamic frames  $DY = \{d_{i}: i=1,\dots,n\}$, and corresponding static frames $ST = \{s_{i}:i=1,\dots,n \}$, along with their respective binary segmentation mask $DY_{seg} = \{d_{i_{seg}}: i=1,\dots,n\}$ and $ST_{seg} = \{s_{i_{seg }}: i=1,\dots,n\}$. The masks consist of two broad classes - static and dynamic. Our goal is to find a mapping from a point on the latent manifold of the static LiDAR scans ($M_s$) to the latent manifold of the dynamic LiDAR scans ($M_d$). Our main challenge is to ensure that the point injections are distributed across a LiDAR in a way that leads to significant navigation deterioration, as compared against a naive random distribution of points.

\subsection{Methodology}
Our model consists of two modules, i.e. a novel segmentation-aware attention autoencoder backbone, $AE_{mask}$ and a pretext-task-based discriminator, $PD$. These are combined in an adversarial setting for point injections \textit{$PiJ$}. \\
$AE_{mask}$ is an autoencoder which utilizes segmentation-based attention coupled with contrastive learning (using hard negatives) to generate a precise reconstruction of the input LiDAR scan. 
$PD$ is trained to discriminate homogeneous and heterogeneous pairs of LiDAR scans. Both these modules are combined for dynamic PiJ in LiDAR scans.\\

\textbf{Segmentation-aware attention based Autoencoder Backbone} - 
\label{ae} 
In this section, we discuss the design of our segmentation-prior-based LiDAR generator, contrastive learning strategies, and Hard Negative mining for contrastive learning on the generator.

For our LiDAR autoencoder backbone ($AE_{mask}$) (Figure \ref{fig:aemask}) we use the Encoder ($H_\phi$) and Decoder ($G_\theta$) from Caccia et al.\cite{caccia2018deep}. We observe that standard LiDAR autoencoder backbones fail to reconstruct the sharp details that are introduced by dynamic objects (examples in \href{https://drive.google.com/drive/folders/1BtrTwUbgFKYkIQ6UGitQAuClHH-pHjBH?usp=sharing}{Supplementary}). Unlike static structures, dynamic structures are not consistent across contiguous LiDAR scans. These inconsistent variations are difficult to learn. We utilize \textit{binary segmentation} mask of LiDAR (stationary v/s non-stationary points) that induces a prior on the reconstructed LiDAR \textit{w.r.t.} the dynamic objects. The binary mask provides explicit attention to the dynamic point features. This is achieved using a segmentation encoder $H_{seg}$. Adaptive average pooling over the hidden layer features of $H_{seg}$ provides channel-level attention to the hidden layer features of $H_\phi$ (encoder of $AE_{mask}$). \\
Given $x\in\{\textit{ST, DY}\}$ and  $x_{seg}\in\{ST_{seg},DY_{seg}\}$ is the corresponding segmentation mask for $x$, our autoencoder, $AE_{mask}$ is defined as follows -
\begin{equation}
     AE_{mask} : (x,x_{seg})\xrightarrow{H_{\phi},H_{seg}} r(x) \xrightarrow{G_{\theta}} \overline{x}
     \quad 
\end{equation}
\textbf{Contrastive Learning on $AE_{mask}$ using Hard negatives}
\label{contrastive}
Static and dynamic LiDAR scans share similar characteristics and structures, but they also have distinct features that aid in better reconstructions. Our experiments show that generative models struggle with regions varying across LiDAR scans. Static objects, with consistent structures across contiguous scans, are easier to learn. In contrast, dynamic structures and occlusions, which vary significantly even across contiguous scans, are more challenging to reconstruct.

LiDAR scans captured in different environments show significant variance. This variance can be leveraged using contrastive learning to learn rich latent representations. We consider 2 approaches for contrasting LiDAR scans - (\textbf{1)} contrast static and dynamic scan (\textbf{2)} contrast scans between different environments (different sequences). We use the former when static-dynamic correspondence is available - CARLA-64, ARD-16 datasets, and the latter when such correspondence is not available (e.g. KITTI). We now describe our method for contrastive learning for LiDAR scans.\\
\begin{figure}[htbp]
\centering
\includegraphics[width=0.8\linewidth, height=0.6\linewidth]{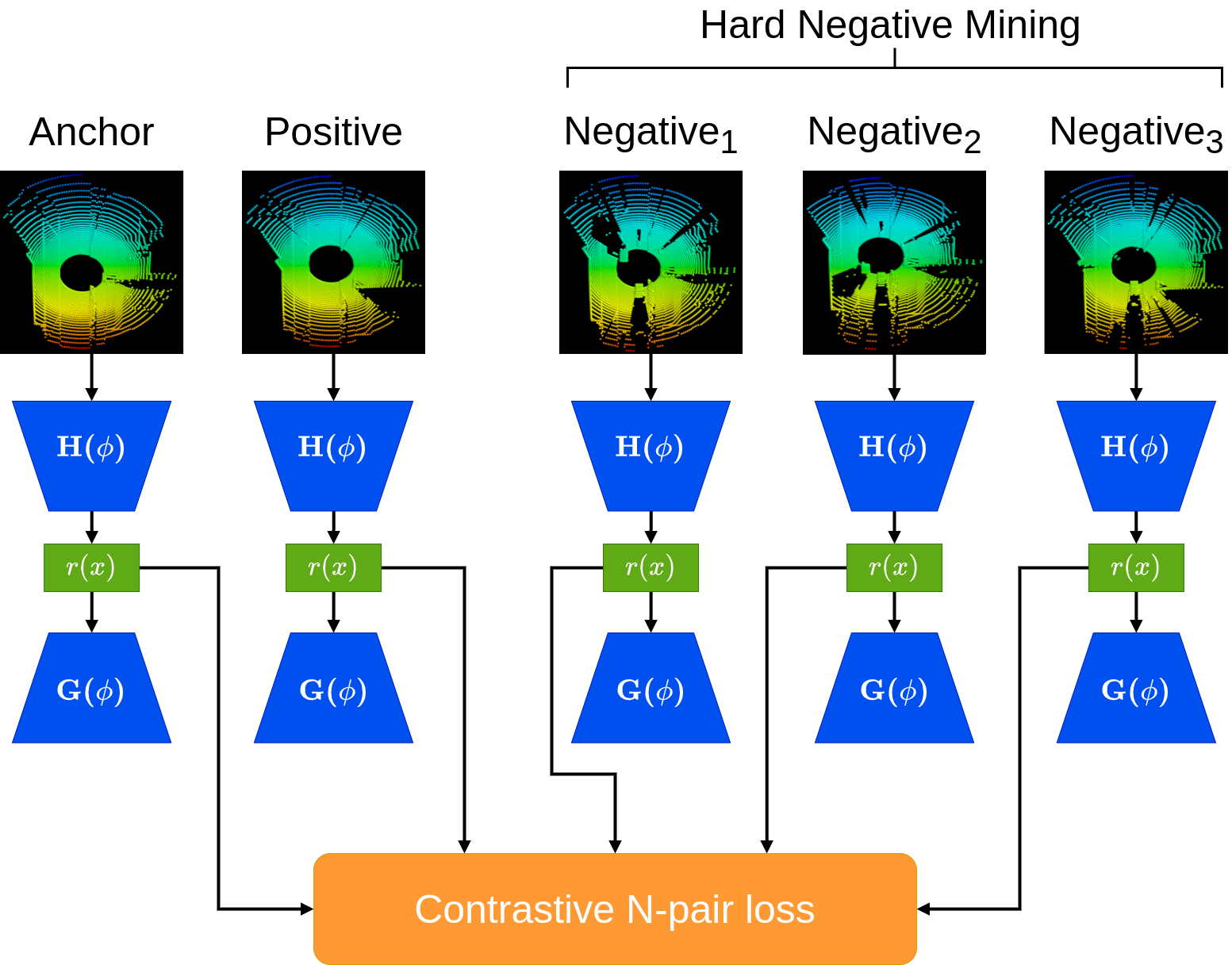}
\caption{Contrastive Loss using Anchor, positive and corresponding dynamic hard negatives.}
\label{fig:contrastive-nppair}
\end{figure}

$\bullet$ Contrasting LiDAR between different Runs :- Several datasets (e.g. KITTI) has unpaired scans from different places (city,
 highway). These sequences can be contrasted against each other to learn better representations. We experimentally validate the effectiveness of this approach.

$\bullet$ Contrast Static with Dynamic scans :- For datasets with correspondence information available - CARLA-64 and ARD-16, we show that exploiting the contrast between these enables precise reconstruction.\\

We describe the losses used for contrasting static-dynamic pairs as well as contrasting different runs of LiDAR scans:

Triplet Loss: It focuses on static-dynamic LiDAR pairs. It pushes similar static scans (anchor \& positive) closer in a latent space, while maximizing the distance between the anchor static scan and its corresponding dynamic scan (negative) with moving objects.

For datasets without static-dynamic correspondence, the anchor and positive sample belong to a particular sequence and the negative samples belong to another sequence. 

N-pair Loss :- It is a generalization of the triplet loss. Instead of using one negative, we use multiple negative samples and contrast the positive sample against them. Experimentally, we find that it leads to better results compared to the triplet loss when used with $AE_{mask}$. For more details, please refer to the  Ablation studies in the \href{https://drive.google.com/drive/folders/1BtrTwUbgFKYkIQ6UGitQAuClHH-pHjBH?usp=sharing}{Supplementary}.


\begin{figure}[h]
\centering
\includegraphics[width=0.6\linewidth, height=0.3\linewidth]{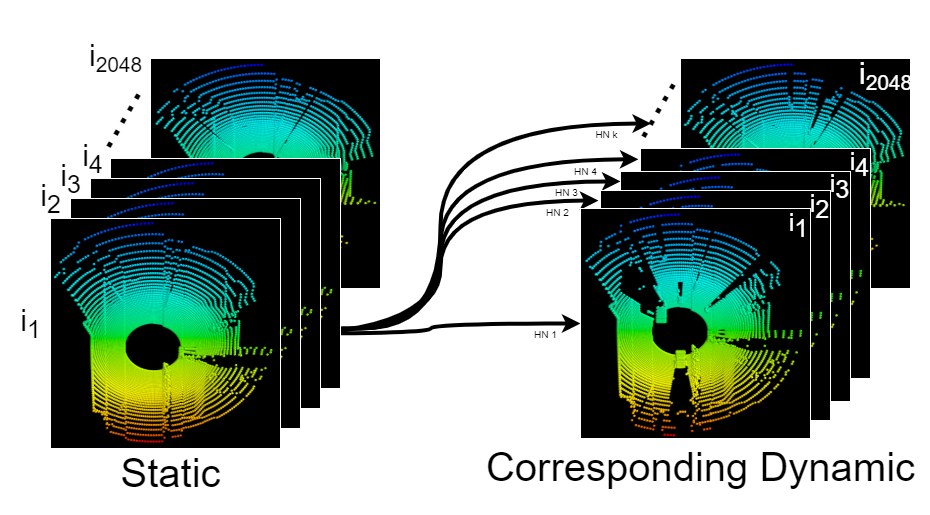}
\caption{Hard Negative mining using corresponding static-dynamic pairs. Given an anchor (static), multiple hard negatives - corresponding dynamic and close-by scans (right) are selected as hard negatives.}
\label{fig:contrastive-hardnegative}
\end{figure}

\begin{figure}[htbp]
\begin{subfigure}{.49\textwidth}
  \centering
 \includegraphics[width=\linewidth]{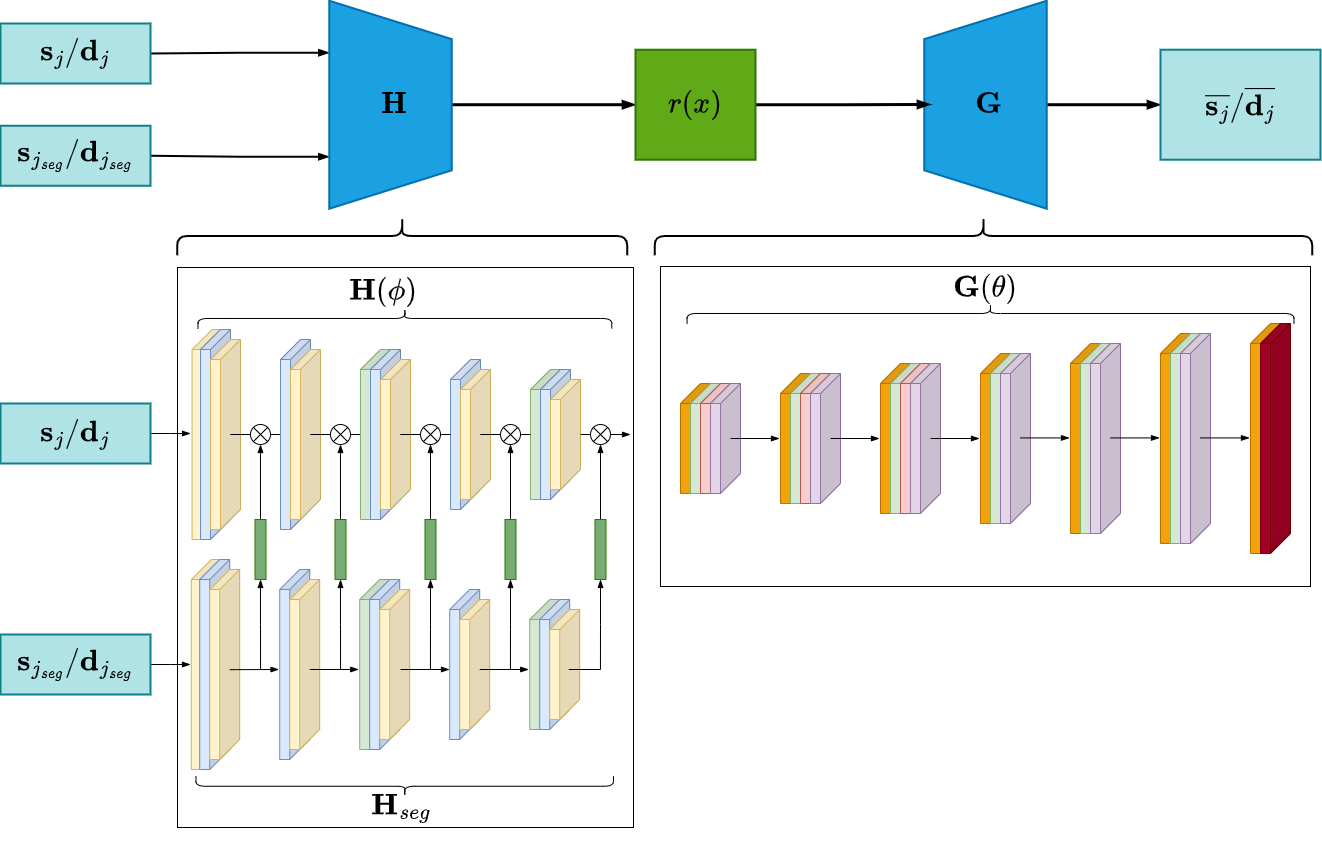} 
  \caption{Segmentation Attention setup for Autoencoder backbone-$AE_{mask}$}
  \label{fig:aemask}
  \end{subfigure}
  \setlength{\tabcolsep}{-5pt}
  \begin{tabular}{c c}
      \includegraphics[width=0.55\linewidth]{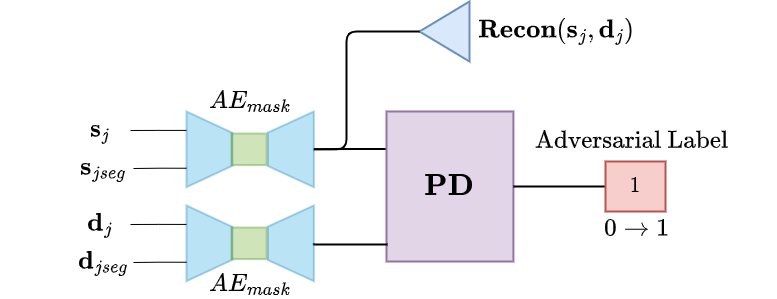} \label{fig:adv}& \includegraphics[width=0.55\linewidth]{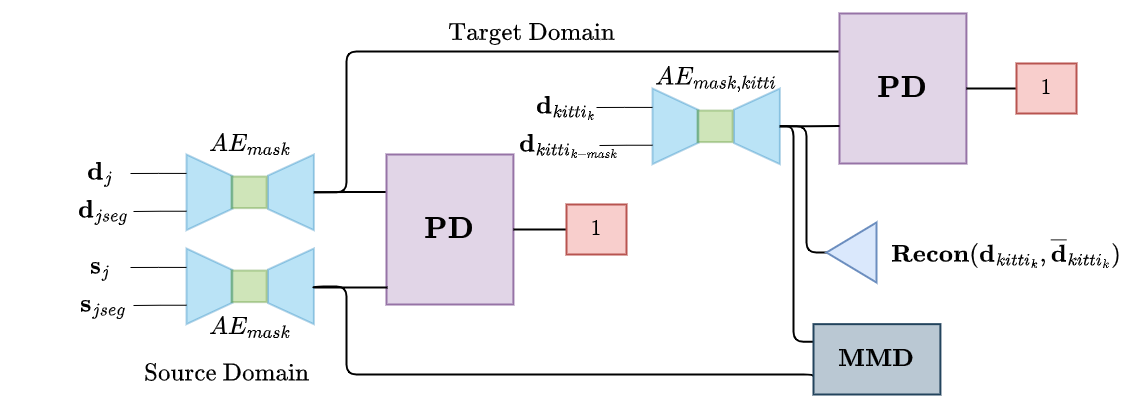} \label{fig:mmd}\\
      \footnotesize{(b) Adversarial Module} & \footnotesize{(c) SLACK-MMD}
      \label{fig:model}
  \end{tabular}

\caption{ \textbf{(a)} Segmentation-based Attention Setup for $AE{mask}$ \textbf{(b)} Adversarial Module tricks PD with an adversarial label. To classify the input pair as 1 - PD injects dynamism in $s_j$. \textbf{(c)} SLACK-MMD adapts CARLA-64 to KITTI.}
\label{fig:arch1}
\end{figure}
Hard-Negative Mining for LiDAR :-
\label{HNM} Contrastive learning performs better with hard negatives—samples close to the anchor in metric space but with different labels. Using dynamic scans as hard negatives for static LiDAR anchors works well, as they share similar structures but differ in dynamic objects and occlusions.

\textbf{Pretext-Task-Discriminator based Adversarial Module} -  Pretext Task training refers to training a model for a \textit{pseudo-task} that helps the model excel at the primary task. We design a discriminator, $PD$ that uses pretext task-based features of LiDAR which help the main task. The pretext task is as follows - latent representations of dynamic scan pairs  ($r_{d_{i}}, r_{d_{j}}$) are given a label '1' (and) heterogeneous pair consisting of a corresponding dynamic and a static scan - ($r_{d_{j}}, r_{s_{j}}$) is given a class '0'. $PD$ contrasts between similarities in dynamic pairs \textit{v/s} difference in static-dynamic pairs. This enables $PD$ to explicitly focus on probable locations in static scans that have dynamic objects in a corresponding dynamic scan. This in turn helps our adversarial module during dynamic point injections.  

The input to $PD$ are latent vectors generated by the autoencoder backbone, $AE_{mask}$. These implicitly encode segmentation-based information in the latent representations. Let $d_i, d_j \in DY$ and $s_j \in ST$ be the corresponding static for $d_i$. Let $r_{d_i}, r_{d_j}, r_{s_j}$ be the corresponding latent representation obtained using $AE_{mask}$ for the above LiDAR scans. $L_{mse}$ is the mean squared error loss while $L_{bce}$ is the Binary cross entropy loss. The final learning objective of $PD$ is
\begin{equation}
\begin{split}
    &=L_{mse}(d_{i},\overline{d_{i}}) + L_{mse}(d_{j},\overline{d_{j}}) +  L_{mse}(s_{j},\overline{s_{j}}) \\
    &+ L_{bce}(PD(r_{d_i} , r_{d_j}), 1) + L_{bce}(PD(r_{d_j}, r_{s_j}), 0)
\end{split}
\label{DILoss}
\end{equation}
 \textit{Adversarial Module} - The adversarial module (Figure 4b) exploits our discriminator, \textit{PD} to attack LiDAR scans with PiJ. We now explain the adversarial module -

The adversarial module assumes that the constituents of the LiDAR pair given to it as input is always of a single type of data (dynamic - 'd'). However, as part of our adversarial trick it is presented with mixed data pairs as input (denoted as 
($d_{j}, s_{j}$)), where 'd' represents dynamic LiDAR data and 's' represents static LiDAR data of the same scene.  During training, the expected label for these mixed pairs is changed from 0 $\rightarrow$ to 1.
This essentially fools the module into treating the static LiDAR data as if it were dynamic. As a result, the adversarial module learns to convert the static LiDAR data (represented in its latent space) into a representation that mimics a LiDAR with dynamic injected points. It results in dynamic point injection $(PiJ)$ in the input LiDAR.


Let ($d_i$, $d_j$, $s_j$) be LiDAR scans involved in the formation of the homogeneous and heterogeneous pairs, where $d_j$ is the corresponding dynamic scan for $s_j$. The adversarial loss  $Adv_{obj}$ is defined as 
\begin{equation}
\begin{split}
    = L_{bce}(PD(r_{d_{j}}, r_{s_{j}}), 1)
    + L_{mse}(s_{j},{d_{j}}) 
\end{split}
\label{DILoss}
\end{equation}

Here, 1 serves as the adversarial label for the heterogeneous pair, which facilitates dynamic PiJ in $s_j$.
\subsection{SLACK-MMD for real-world settings}

There exists a challenge in applying our method to real-world datasets like KITTI because they often lack corresponding pairs of dynamic and static LiDAR scans. Additionally, models trained on datasets with these pairs might not perform well on KITTI due to differences between the data sources (domain shift). To overcome this hurdle and ensure our method works seamlessly on datasets like KITTI, we modify our adversarial module using a technique called unsupervised domain adaptation (UDA).

We minimize the domain distance between the source and the target domain in the latent space (Figure 4c). We use the Maximum Mean Discrepancy (MMD) loss from Borgwardt et. al \cite{Borgwardt2006IntegratingSB} to maximize the domain invariance. There exist several methods in the literature to minimize the discrepancy between latent vectors in different domains. We use \cite{Borgwardt2006IntegratingSB} because it is simple, easy to use, and works well in our settings.\\
We initialize the UDA network with weights of the autoencoder backbone, $AE_{mask}$ and the discriminator $PD$ that is obtained after the adversarial training for CARLA-64.
 The autoencoder responsible for attacking KITTI scans - $AE_{mask_{kitti}}$, is pre-trained separately using segmentation-based attention and contrastive leaning. Using a separate KITTI autoencoder ensures that dynamic injections are explicitly done on the KITTI latent manifold. 

Latent representations of the dynamic source scan ($d_j$) and target scan ($d_{kitti_j}$) are used to calculate the discrepancy between the domains. These latent representations are also fed to discriminator \textit{PD} with an adversarial label of 1. It ensures that backpropagation injects dynamism in $d_{kitti_j}$.

Given LiDAR scan pair - $d_j$, $s_j$ and KITTI scan $d_{kitti_j}$, training loss for KITTI dataset $Loss _{MMD_{kitti}}$ is
\begin{equation}
\begin{split}
    &=L_{bce}(PD(r_{d_{j}}, r_{s_{j}}), 1) 
    + L_{bce}(PD(r_{d_{j}}, r_{d_{kitti_j}}), 1)\\
    &+ L_{mse}(d_{kitti_j},\overline{{d_{kitti_j}}})
\end{split}
\label{UDA}
\end{equation}

\section{Experiments}
\label{sec:exp}
Our experiments are divided into 3 parts - \textbf{(a)} We evaluate our autoencoder backbone, $AE_{mask}$ against standard LiDAR autoencoder backbones to show the benefit of segmentation-assisted attention and contrastive learning,
\textbf{(b)} We evaluate SLACK on simulated and real-world LiDAR datasets for PiJ, and
\textbf{(c)} We also evaluate the impact of PiJ attacks using SLACK on navigation using SLAM. Henceforward we divide the experiments and evaluation in these 3 parts as above. \\

\textbf{Datasets} - We use 3 datasets to test SLACK - CARLA-64 \cite{kumar2021dslr}, KITTI Odometry dataset \cite{geiger2012we}, and ARD-16 \cite{kumar2021dslr}. We provide more details on these in the \href{https://drive.google.com/drive/folders/1BtrTwUbgFKYkIQ6UGitQAuClHH-pHjBH?usp=sharing}{Supplementary}.\\
\\
\textbf{Evaluation Metrics}

$AE_{mask}$ - To evalaute our autoencoder backbone $AE_{mask}$ ,  we use two standard metrics - Earth Mover's and Chamfer Distance \cite{emd-cd}.\textbf{SLACK} - To evaluate the quality of dynamic scans generated by SLACK we use 2 baselines. - LiDAR Quality Index (LQI) and Dynamic Segmentation Ratio (DSR).
LQI is used to assess the quality of a given  LiDAR scan. It regresses the amount of noise in a given LiDAR scan and is based on the CNN IQA model \cite{kang2014convolutional}. It is based on the assumption that dynamic objects are noise in the LiDAR distribution, with noise level estimating quality and dynamism. DSR quantifies the percentage of dynamic points, using a network trained to classify LiDAR points as dynamic or static, providing a per-point binary segmentation. For more details on these please refer to \href{https://drive.google.com/drive/folders/1BtrTwUbgFKYkIQ6UGitQAuClHH-pHjBH?usp=sharing}{Supplementary}.

\textbf{Note}: \textit{A viable attack model requires the attacked LiDAR scan to have low LQI and high DSR. High LQI indicates detectable deviations from the original, while low DSR indicates a failure to insert new dynamism.}\\
\label{howtoreadnumbers}
\textbf{Effect of SLACK on SLAM} - 
To evaluate PiJ attacks of SLACK on SLAM, we use Google Cartographer\cite{hess2016real}, a LiDAR-based SLAM algorithm. We use two metrics for translation and rotational error induced by SLACK - Absolute Trajectory Error (ATE)\cite{sturm2012benchmark} and Relative Pose Error (RPE) \cite{sturm2012benchmark}. For details on these metrics, please refer to \href{https://drive.google.com/drive/folders/1BtrTwUbgFKYkIQ6UGitQAuClHH-pHjBH?usp=sharing}{Supplementary}.

\textbf{Baselines}
\label{bl-ae-mask}
\textbf{$AE_{mask}$} -
We evaluate our LiDAR autoencoder backbone, $AE_{mask}$ against backbone architectures that have been used successfully for LiDAR generative modeling. We compare $AE_{mask}$ with methods that work in real-time and do not require additional data in different modalities during training. We select the following based on criteria: \CP \cite{xu2023cp3}, \atlas{}\cite{groueix2018papier}, \Ach{}\cite{achlioptas2017representation}, \CacciaAE{}, and \CacciaVAE{}, \cite{caccia2018deep}. For details on baselines, please refer to \href{https://drive.google.com/drive/folders/1BtrTwUbgFKYkIQ6UGitQAuClHH-pHjBH?usp=sharing}{Supplementary}.
\textbf{SLACK} - We evaluate dynamic point injections (PiJ) using SLACK against several baselines. Criteria is that baseline must work in real-time without the need of data in other modalities we adopt the following models for PiJ for comparison - \Ach{}  \cite{achlioptas2017representation},  \CacciaAE{},  \CacciaVAE{},  \CacciaGAN{} \cite{caccia2018deep}, and \DSLR{}\cite{kumar2021dslr}.\begin{table}[t]
\centering
\scriptsize
\begin{tabular}{l|ll|ll|ll}
\hline
Model & \multicolumn{2}{c|}{CARLA-64} & \multicolumn{2}{c|}{KITTI} & \multicolumn{2}{c}{ARD-16} \\ \hline
 & \multicolumn{1}{c}{Chamfer}&EMD& \multicolumn{1}{c}{Chamfer} & EMD&\multicolumn{1}{c}{Chamfer}&EMD    \\ \hline
\atlas{} & \multicolumn{1}{c}{11.56} & 1208 & \multicolumn{1}{c}{2.85} & 1571 & \multicolumn{1}{c}{3.53} & 392.4 \\ 
 \Ach{}& \multicolumn{1}{c}{1.91} & 696 & \multicolumn{1}{c}{2.16} & 1103 & \multicolumn{1}{c}{0.62} & 290.7 \\ 
\CacciaVAE{} & \multicolumn{1}{c}{1.82} & 157 & \multicolumn{1}{c}{1.16} & 144 & \multicolumn{1}{c}{0.33} & 72.0 \\ 
\CacciaAE{} & \multicolumn{1}{c}{1.52} & 164 & \multicolumn{1}{c}{0.65} & 141 & \multicolumn{1}{c}{0.33} & 63.8 \\
$AE_{mask}$\textbf{(Ours)} & \multicolumn{1}{c}{\textbf{0.93}} & \textbf{126} & \multicolumn{1}{c}{\textbf{0.57}} & \textbf{130} & \multicolumn{1}{c}{\textbf{0.30}} & \textbf{63.3} \\ \hline
\end{tabular}
\caption{Comparison of our autoencoder backbone - $AE_{seg}$ with widely used LiDAR backbone autoencoders.}

\label{tab:aeseg}
\end{table}

\begin{table}[]\centering
\scriptsize
\begin{tabular}{l|rr|rr|r}\hline
&\multicolumn{2}{c}{KITTI-64}  &\multicolumn{2}{c}{CARLA-64} &ARD-16 \\ \hline
\multicolumn{1}{c}{Model} &\multicolumn{1}{c}{LQI$\downarrow$} &\multicolumn{1}{c}{DSR$\uparrow$} &\multicolumn{1}{c}{LQI$\downarrow$} &\multicolumn{1}{c}{DSR$\uparrow$} &\multicolumn{1}{c}{LQI} \\\hline
\CP{} &\textbf{0.52} &\multicolumn{1}{c|}{\red{0.16}} &1.54 &\multicolumn{1}{c|}{\red{0.31}} &- \\
\Ach{} & \red{5.95} &\multicolumn{1}{c|}{\textbf{0.48}} & \red{7.64} &\multicolumn{1}{c|}{\textbf{0.62}} &- \\
\CacciaAE{} &3.28 &\multicolumn{1}{c|}{0.44} &3.93 &\multicolumn{1}{c|}{0.47} &0.58 \\
\CacciaVAE{} &3.4 &\multicolumn{1}{c|}{0.44} &4.32 &\multicolumn{1}{c|}{0.49} &0.61 \\
\CacciaGAN{} &3.84 &\multicolumn{1}{c|}{0.43} &5.47 &\multicolumn{1}{c|}{0.50} &\textbf{0.43} \\
\DSLR{} &3.32 &\multicolumn{1}{c|}{0.43} &4.41 &\multicolumn{1}{c|}{0.48} &- \\
\SAT &\textbf{1.97} &\multicolumn{1}{c|}{\textbf{0.48}} &\textbf{3.73} &\multicolumn{1}{c|}{\textbf{0.51}} &0.68 \\
\hline
\end{tabular}
\caption{Comparison of SLACK with baselines for PiJ attacks. Red indicates values that are bad and cannot be used for PiJ - very high LQI or very low DSR. Please refer to the Note in Section \ref{sec:exp} for interpreting the numbers. A method needs to perform well on both metrics to be usable for attacks. We do not report DSR for ARD-16 as it does not have segmentation details. SLACK does not work well with ARD-16.}
\label{table:slacknum}
\end{table}

\textbf{Effect of SLACK on SLAM} -
We the impact of a PiJ attack on SLAM. To ensure a fair comparison, we define criteria for baseline attacks: \textbf{(1)Similar LiDAR quality} Attacked scans should have quality equal to or better than a benchmark (SLACK) and \textbf{(2)Sufficient attack points} The number of points attacked by the baseline must be at least as many as those attacked by SLACK. These ensure that the attack injects sufficient adversarial points to destabilize navigation without ruining the LiDAR quality. 
 We propose two baselines:\textbf{(a)} Random Point Removal (\RR{}) - we randomly remove \textit{k}  points from the original LiDAR -  \textit{k} being the number of new point injections by SLACK. The point removal strategy follows a Bernoulli distribution with the mean equal to the percentage of points attacked by SLACK. \textbf{(b)} Random Point Injection (\RN{}) - we randomly inject noise into  \textit{k} points in the original LiDAR scan. The magnitude of the noise injected in the \textit{k} random points is the same as the magnitude of the new dynamic injections in the LiDAR scan attacked by SLACK. 

We choose these baselines instead of the baselines in Table \ref{table:slacknum} due to the following reason - the focus of our work is to attack LiDAR without deteriorating LiDAR quality. These baselines in Table \ref{table:slacknum} fail to retain adequate LQI and can be identified from the original unattacked LiDAR.

\subsection{Results}

\label{sec:res}
\textbf{LiDAR Autoencoder Backbone} - We compare $AE_{mask}$ against the backbone baselines discussed in Section \ref{bl-ae-mask} in Table \ref{tab:aeseg}. Our proposed model performs better on both metrics across the baselines. The segmentation-assisted attention and the contrastive learning helps $AE_{mask}$ to learn better representations of dynamic regions and reconstructs them precisely compared to the baselines (figures in \href{https://drive.google.com/drive/folders/1BtrTwUbgFKYkIQ6UGitQAuClHH-pHjBH?usp=sharing}{Supplementary}). We show the effect of the segmentation and the contrastive module in the Ablation studies in the \href{https://drive.google.com/drive/folders/1BtrTwUbgFKYkIQ6UGitQAuClHH-pHjBH?usp=sharing}{Supplementary}.

\begin{figure}
\begin{tabular}{cc}
    \includegraphics[width=0.5\linewidth]{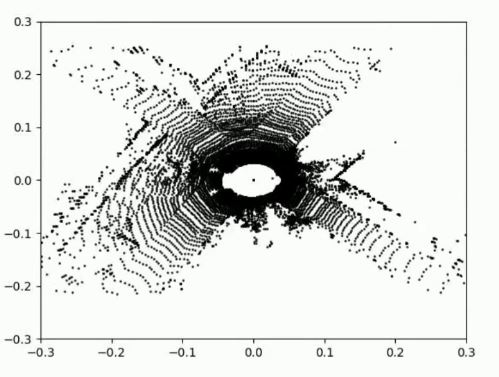} & \includegraphics[width=0.5\linewidth]{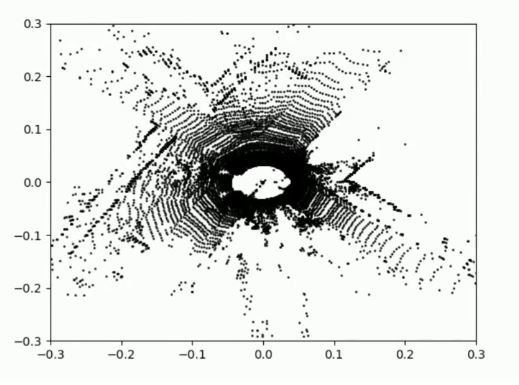} \\
    Original Scan & Attacked Scan
\end{tabular}
\caption{Visual demonstration of an original scan v/s an attacked scan. It is very difficult to distinguish the attacked scan from the original scan. Video demo in the  \href{https://drive.google.com/drive/folders/1BtrTwUbgFKYkIQ6UGitQAuClHH-pHjBH?usp=sharing}{Supplementary}.}
\label{fig:kitti-attack}
\end{figure}

\begin{table}[t]
\centering
\scriptsize
\begin{tabular}{lrrrrrr}\toprule
\makecell[l]{KITTI\\Seq}&\makecell[l]{PiJ\\(\%)}& No Attack &\makecell[l]{Rand.\\Rem. (\RR{})} &\makecell[l]{Rand.\\Injection (\RN{})}&\makecell[l]{SLACK\\(Ours)} \\\cmidrule{1-6}
& &ATE/RPE &ATE/RPE &ATE/RPE &ATE/RPE \\\midrule
\multicolumn{6}{c}{KITTI} \\ \hline
0 &0.017 &22.97/\textbf{1.12} &25.92/\textbf{1.12} &25.26/\textbf{1.12} &\textbf{31.68}/1.10 \\

1 &0.068 &415.71/2.32 &600.60/2.62 &342.29/1.93 &\textbf{665.67}/\textbf{3.03} \\

2 &0.06 &139.37/1.76 &153.00/\textbf{1.79} &166.40/1.77 &\textbf{167.188}/1.73 \\

4 &0.065 &103.10/2.57 &87.92/1.86 &108.48/1.77 &\textbf{108.6/3.71} \\

5 &0.07 &7.65/1.22 &7.24/1.21 &13.26/1.21 &\textbf{40.46/1.25} \\

6 &0.08 &5.29/1.64 &127.57/2.18 &\textbf{144.34}/2.38 &142.53/\textbf{2.42} \\

7 &0.076 &4.15/\textbf{1.05} &5.174/1.04 &6.29/1.03 &\textbf{6.55}/1.04 \\

8 &0.071 &196.48/13.30 &\textbf{196.96}/13.33 &191.61/\textbf{13.93} &194.59/13.00 \\

9 &0.07 &11.94/\textbf{1.77} &12.24/1.76 &31.03/1.74 &\textbf{221.66}/1.49 \\

10 &0.14 &6.52/\textbf{1.37} &5.08/1.35 &48.60/1.29 &\textbf{108.98}/1.16 \\ 

\hline
\multicolumn{6}{c}
{CARLA} \\  \hline
1 &0.089 &0.51/\textbf{0.19} &2.69/0.13 &1.72/0.1 &\textbf{3.33}/0.11 \\

2 &0.088 &0.39/0.04 &0.70/0.07 &0.9/0.08 &\textbf{1.38/0.10} \\
\bottomrule
\end{tabular}
\caption{Comparison of PiJ attacks on SLAM. SLACK deteriorates SLAM more than the baselines while using the same number of injected points and maintaining better LIDAR quality. Note that the percentage of points injected by SLACK is determined by the sequence and the model.}
\label{tab:attack-res}
\end{table}

\textbf{SLACK} - For CARLA-64 and KITTI datasets, we observe that \SAT{} generates better quality of injected dynamic LiDAR scans (Table \ref{table:slacknum}). \SAT{} maintains better LQI than most baselines \textbf{and} inserts considerable PiJ points across the LiDAR scan. \textit{\CP{} achieves better LQI than \SAT{}, but the dynamic points in the attacked LiDAR scan are too low (low DSR), making the model unfit for PiJ} Refer Note in Section \ref{sec:exp}). We demonstrate an attacked scan in Figure \ref{fig:kitti-attack}. It is difficult to detect and differentiate the attacked scan from the original scan.
We observe that \textit{\Ach{} fails to maintain good LiDAR quality (high LQI), although it gives better DSR compared to SLACK.} \SAT{} performs well on both metrics as a whole.

\begin{figure}[th]
    \centering    \includegraphics[width=\linewidth]{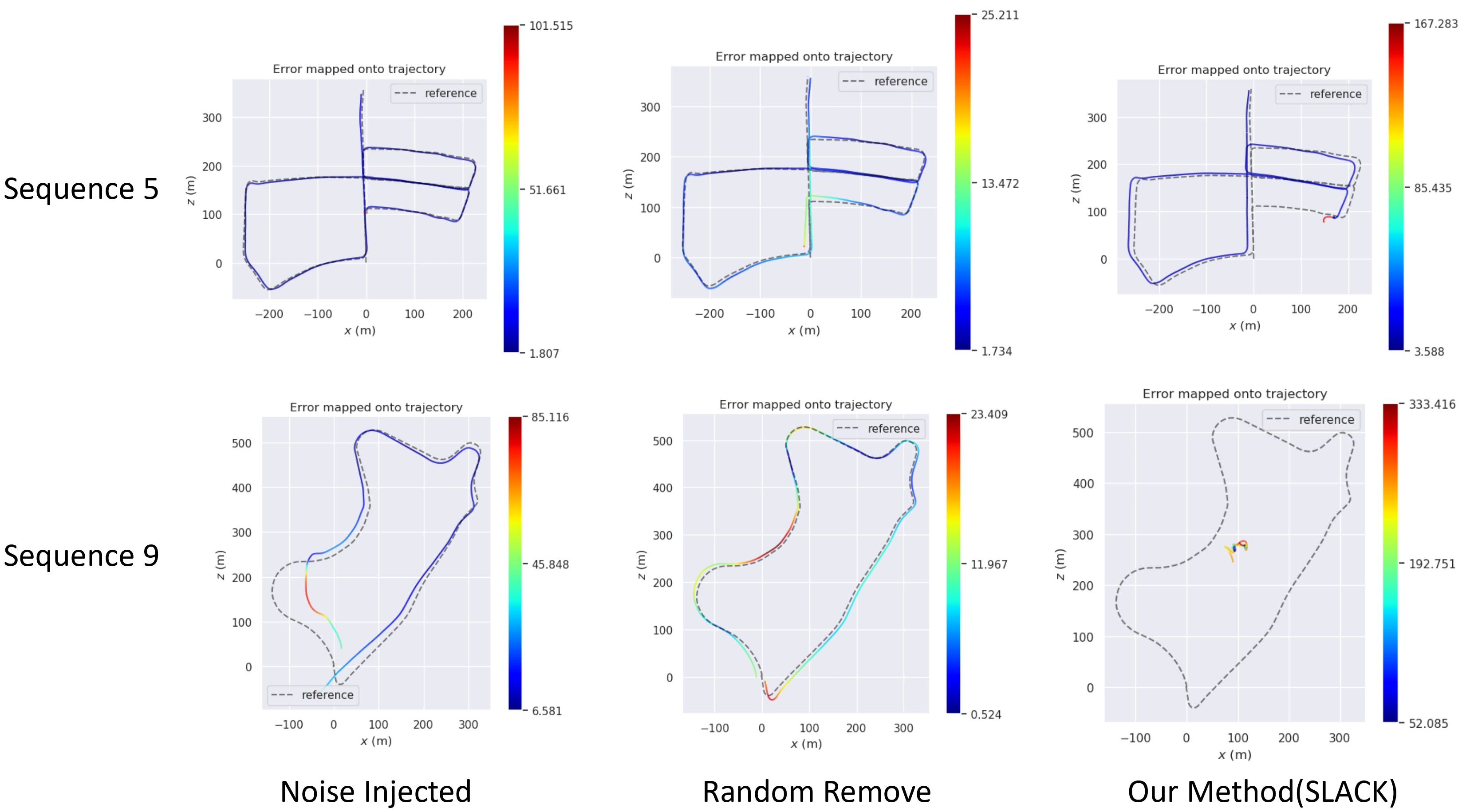}
    \caption{Attack comparison with baselines on KITTI. Dotted line: GT trajectory. Solid line: trajectory after the attack.}
    \label{fig:kitti-slam-attack2}
\end{figure}

For ARD-16 dataset, we evaluate our model on the ARD-16 dataset in Table \ref{table:slacknum}. We observe that our model does not perform well against them. We do not report the Dynamic Segmentation Ratio here as ARD-16 does not have a dynamic segmentation mask available. A strong reason for the poor performance is that ARD-16 it is a 16-beam sparse LiDAR dataset. It has fewer points falling on dynamic objects, which is too low for our model to learn anything. Geometric and hardware-based methods \cite{cao2019adversarial,sun2020towards} may perform better than our model in such scenarios. Our model is not robust against very few beam-based LiDAR point clouds.

\textbf{Effects of SLACK on SLAM} - In this section, we demonstrate the effect of PiJ attacks using SLACK on downstream SLAM performance.

 We provide a quantitative demonstration of the attacked LIDAR sequences for CARLA-64 and KITTI in Table \ref{tab:attack-res} as well as in Figure \ref{fig:slam-first} and \ref{fig:kitti-slam-attack2}. 
For CARLA-64, SLACK shows consistently higher translation error (ATE) than baselines despite maintaining LiDAR quality and an equal number of point injections. Our experiments highlight that the \textit{location of injection matters more than the number of points}, with SLACK strategically injecting points to severely degrade navigation. We qualitatively demonstrate the degraded trajectory and map quality in Figure \ref{fig:slam-first}. 

We also provide a qualitative demonstration of the SLACK PiJ attacks on KITI sequences in Figure \ref{fig:kitti-slam-attack2}
 and \ref{fig:kitti-slam-attack2}. 
 A small number of PiJ is also needed to destabilize the SLAM trajectory, map quality and navigation are severely affected (Figure \ref{fig:kitti-slam-attack2}).  
   Sequences that have minimal loop closures - 0 or 1, e.g. sequence 1,4,6,9,10 have consistently higher errors due to SLACK.  We provide a video demo in the \href{https://drive.google.com/drive/folders/1BtrTwUbgFKYkIQ6UGitQAuClHH-pHjBH?usp=sharing}{Supplementary} to compare an attacked LiDAR vs original. Notice it is impossible to identify the attacked LiDAR.

\section{Analysis and Conclusion}
 This research aims to raise awareness about the criticality of PiJ attacks on LiDAR. By understanding the potential consequences, researchers and developers can focus on implementing robust security measures to ensure safe and reliable operation of AVs'.
 While simulated datasets offer controlled testing, we acknowledge the need for real-world validation. We demonstrate the  impact of SLACK on real LiDAR scans on the KITTI dataset. To bridge the gap between simulation and deployment, future collaborations with car manufacturers for testing on actual LiDAR systems are crucial for understanding an attack's true feasibility and impact. It is assumed that this is a white box attack setting where the attacker has gained full access to the model and LiDAR scanner. The aim of the research is to exhibit the potential for attacks on LiDAR systems. 
 From Table \ref{tab:attack-res}, we conclude that an attack on LiDAR-based SLAM requires a small amount of injected points at strategic locations while preserving the LiDAR quality. This leads us to conclude that navigation accuracy may rely on certain strategic points, which, when destroyed by PiJ, can affect downstream task. Another interesting observation is that sequences with multiple loop closures(2,8) are not affected by SLACK \cite{kumar2024slack,Kumar_2023_BMVC,kumar2024glidr,kumar2024moves}. These sequences may be able to negate the effect of PiJ by using loop closures to reduce overall error. 

\bibliographystyle{IEEEbib}
\bibliography{refs}

\end{document}


%

\newcommand{\atlas}{{\textsc{AtlasNet}}}
\newcommand{\Ach}{{\textsc{Achlioptas et al.}}}

\newcommand{\CacciaAE}{{\textsc{Caccia-AE}}}
\newcommand{\SAT}{{\textsc{SLACK}}}
\newcommand{\CP}{{\textsc{CP3}}}
\newcommand{\SATUDA}{{\textsc{SLACK-MMD}}}
\newcommand{\CacciaVAE}{{\textsc{Caccia-VAE}}}
\newcommand{\CacciaGAN}{{\textsc{Caccia-GAN}}}
\newcommand{\rulesep}{\unskip\ \vrule\ }
\newcommand{\DSLR}{\textsc{DSLR}}
\newcommand{\RR}{\textsc{RR}}
\newcommand{\RN}{\textsc{RN}}
\newcommand{\red}[1]{{#1}}

\maketitle


\maketitle
\begin{figure}
\begin{subfigure}{.50\textwidth}
  \centering
  \includegraphics[width=1\linewidth, height=0.5\linewidth]{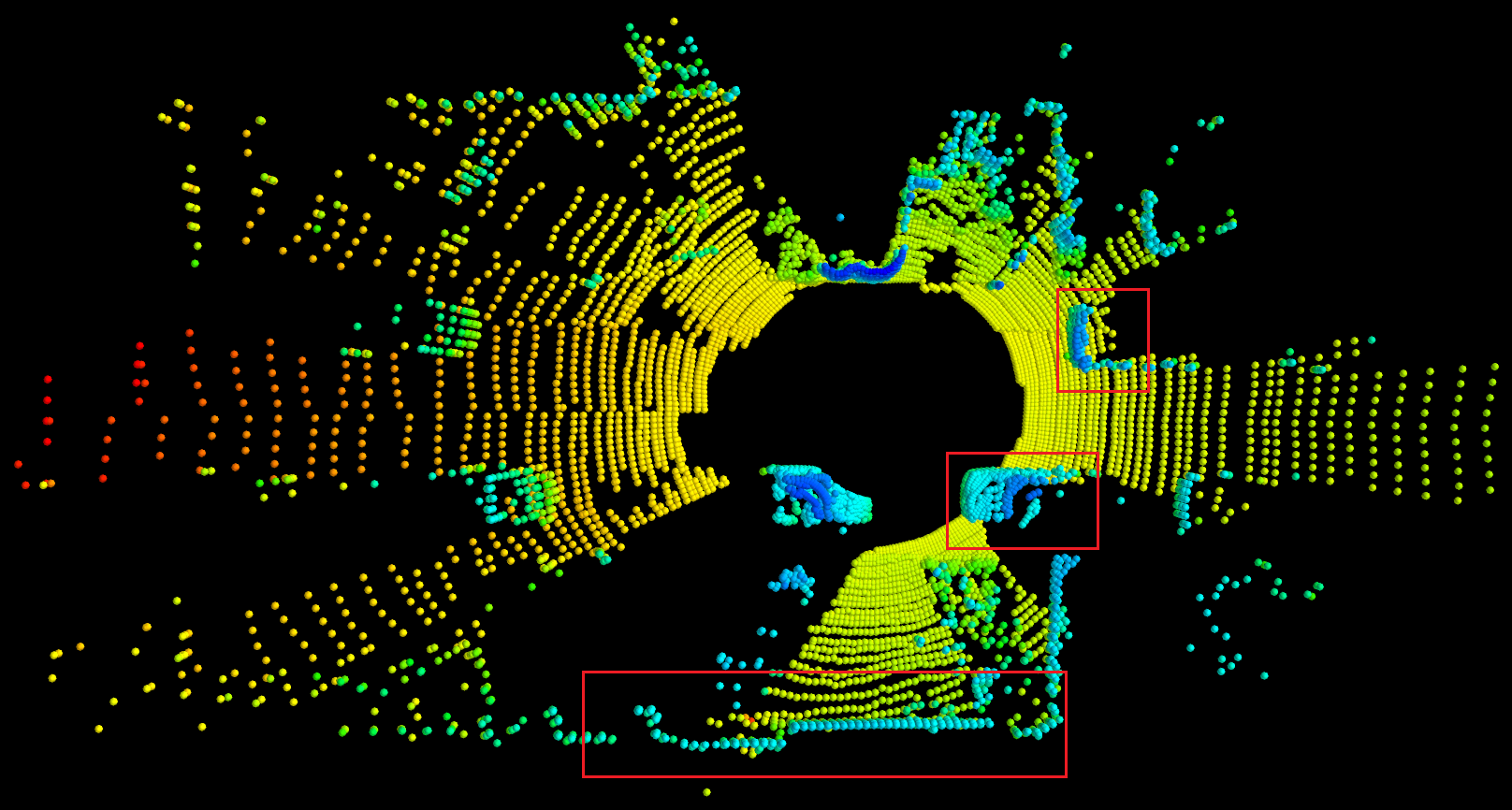}  
  \caption{}
  \label{fig:sub-first}
\end{subfigure}
\begin{subfigure}{.50\textwidth}
  \centering
  \includegraphics[width=1\linewidth, height=0.5\linewidth]{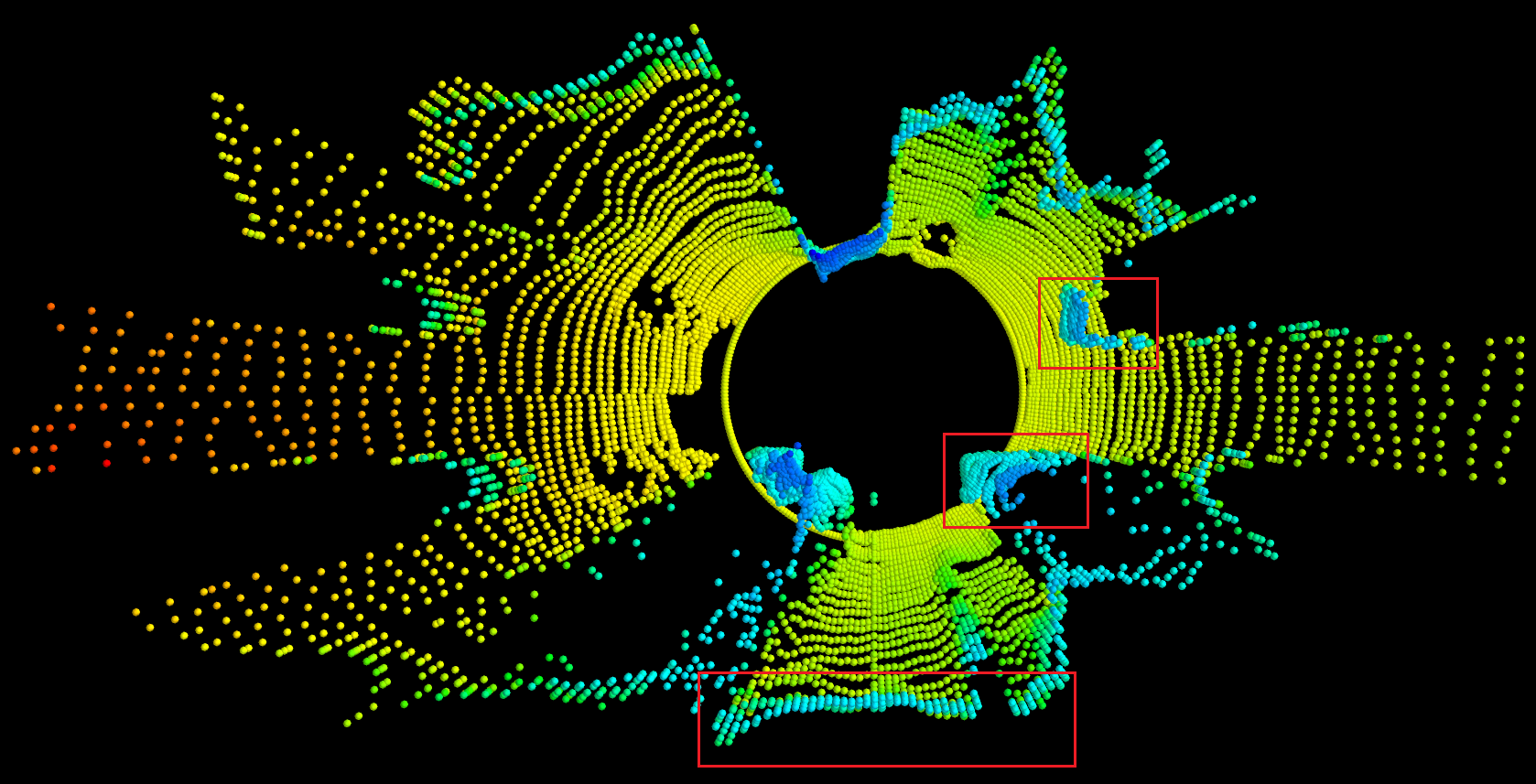}  
  \caption{\textbf{}}
  \label{fig:sub-first}
\end{subfigure}

\begin{subfigure}
{0.5\textwidth}
\centering 

  \centering
  \includegraphics[width=1\linewidth, height=0.5\linewidth]{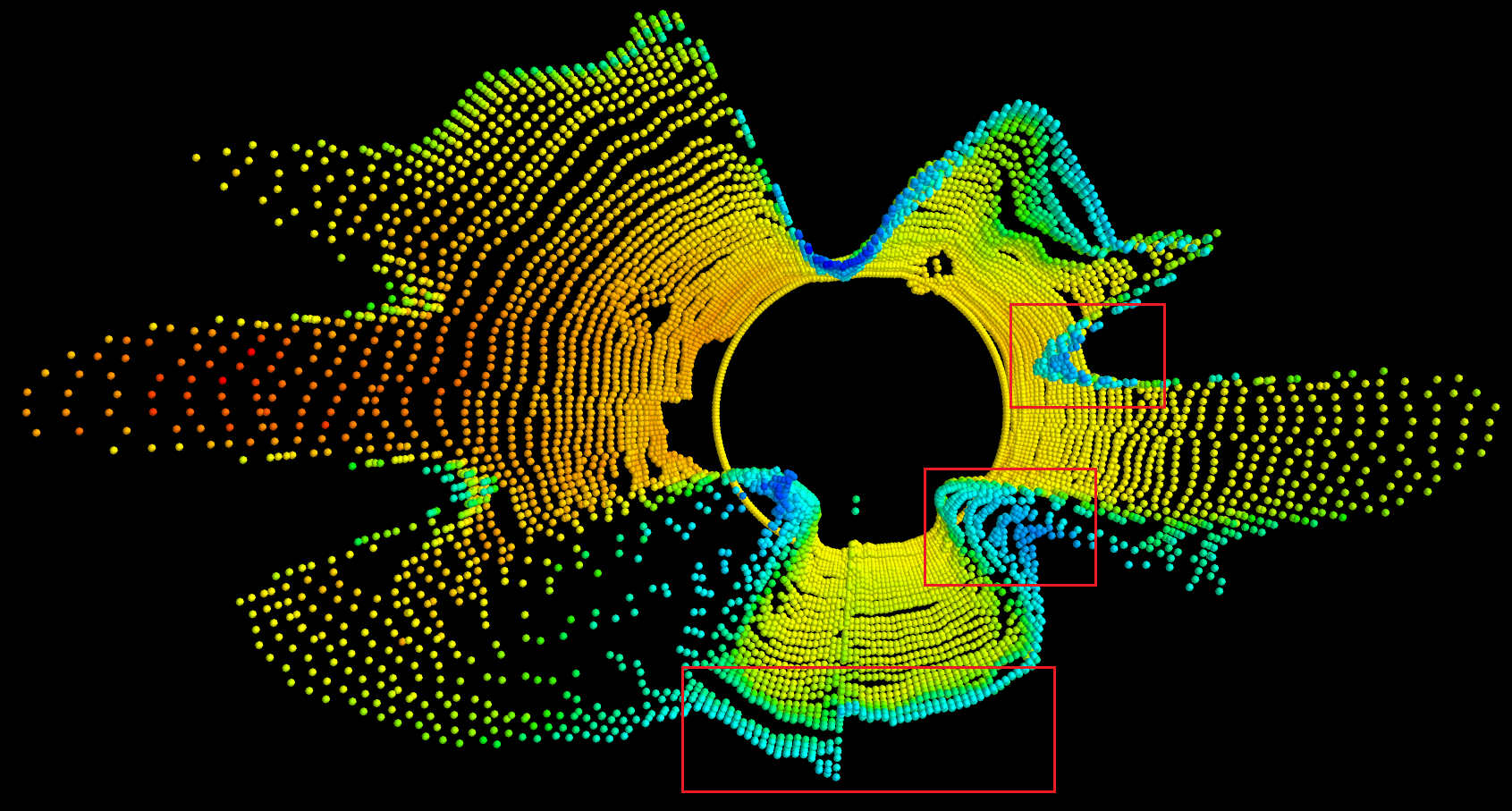}  
  \caption{}
  \label{fig:sub-first}
\end{subfigure}

\caption{Comparison of $AE_{mask}$ results with best baseline in high resolution. \textbf{Top} - original KITTI scan. \textbf{Middle} - reconstruction using $AE_{mask}$ preserves precise static and dynamic structures. \textbf{Bottom} - reconstruction using the best baseline. }
\label{kittiimages}
\end{figure}

\section{Video Demo of SLACK and visualization of $AE_{mask}$ reconstruction}
\begin{figure*}[t]
\begin{subfigure}{.33\textwidth}
  \centering
  \includegraphics[width=1\linewidth, height=0.5\linewidth]{images/original47.png}  
  \caption{Orignal Scan}
  \label{fig:a}
\end{subfigure}
\begin{subfigure}{.33\textwidth}
  \centering
  \includegraphics[width=1\linewidth, height=0.5\linewidth]{images/ours47.png}  
  \caption{Our Autoencoder reconstruction}
  \label{fig:b}
\end{subfigure}
\begin{subfigure}{.33\textwidth}
  \centering
  \includegraphics[width=1\linewidth, height=0.5\linewidth]{images/baseline47.png}  
  \caption{Standard LiDAR autoencoder backbones}
  \label{fig:c}
\end{subfigure}

\begin{subfigure}{.33\textwidth}
  \centering
  \includegraphics[width=1\linewidth, height=0.5\linewidth]{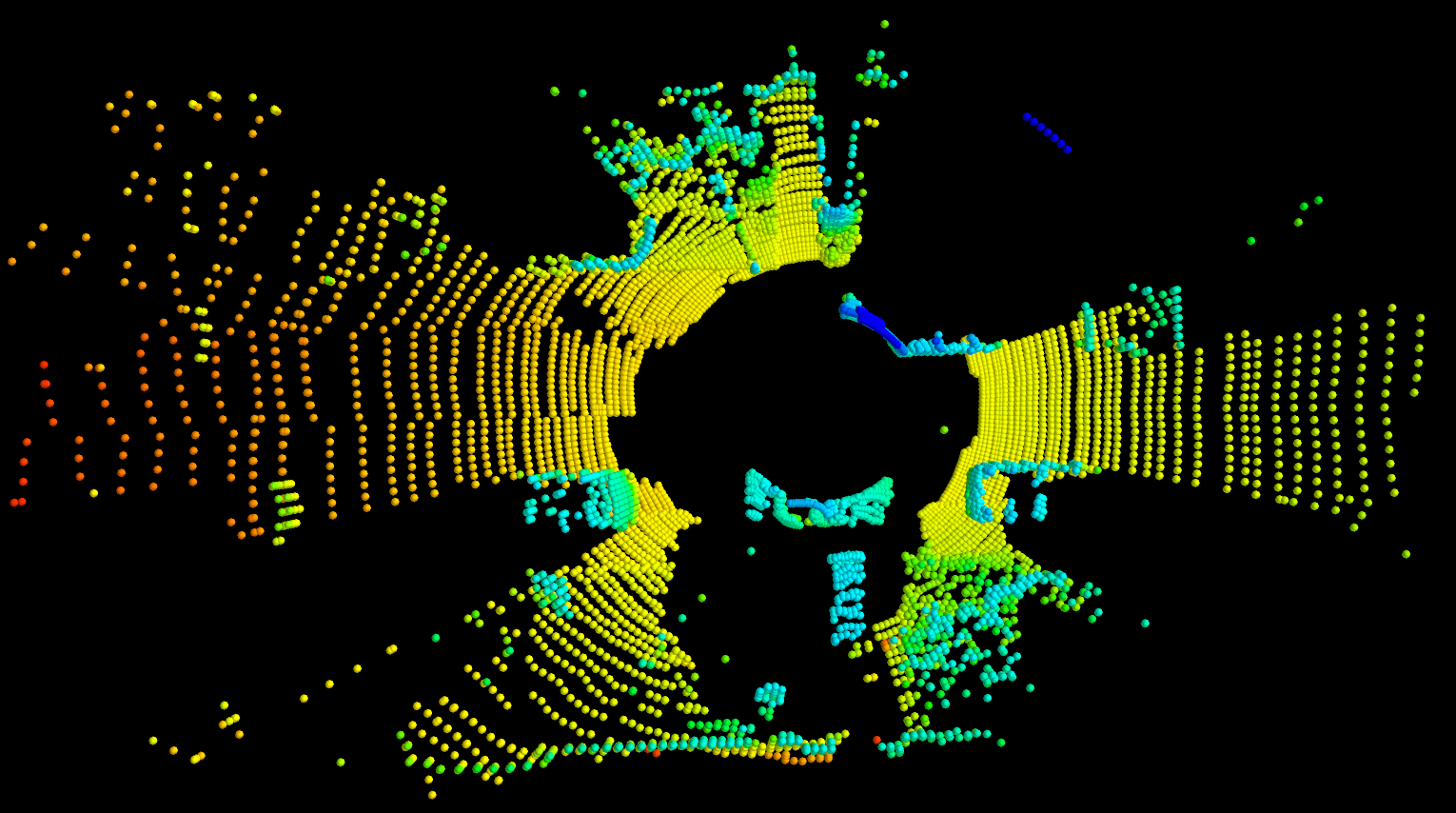}  
  \caption{Original input scan}
  \label{fig:d}
\end{subfigure}
\begin{subfigure}{.33\textwidth}
  \centering
  \includegraphics[width=1\linewidth, height=0.5\linewidth]{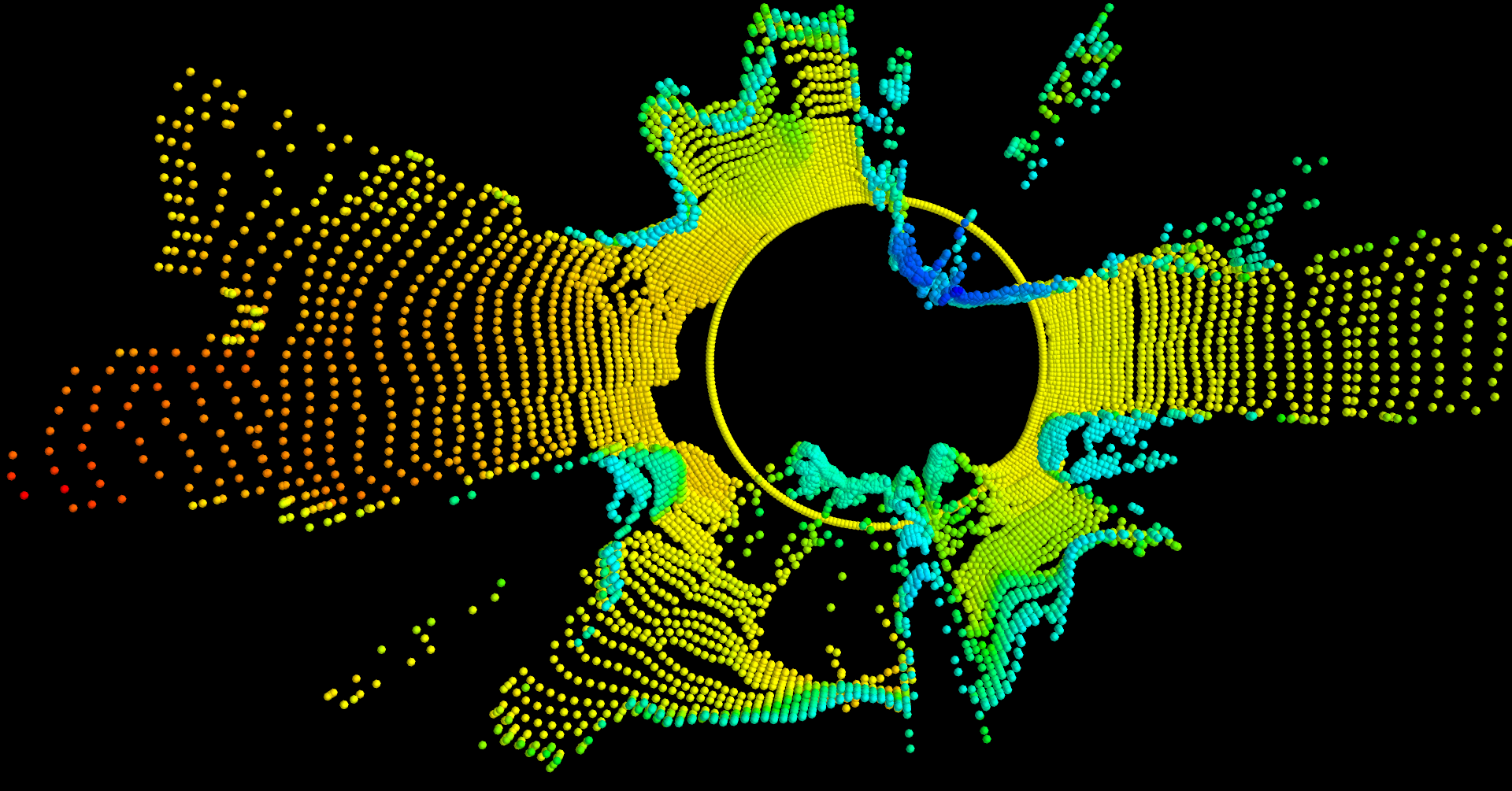}  
  \caption{Our autoencoder reconstruction}
  \label{fig:e}
\end{subfigure}
\begin{subfigure}{.33\textwidth}
  \centering
  \includegraphics[width=1\linewidth, height=0.5\linewidth]{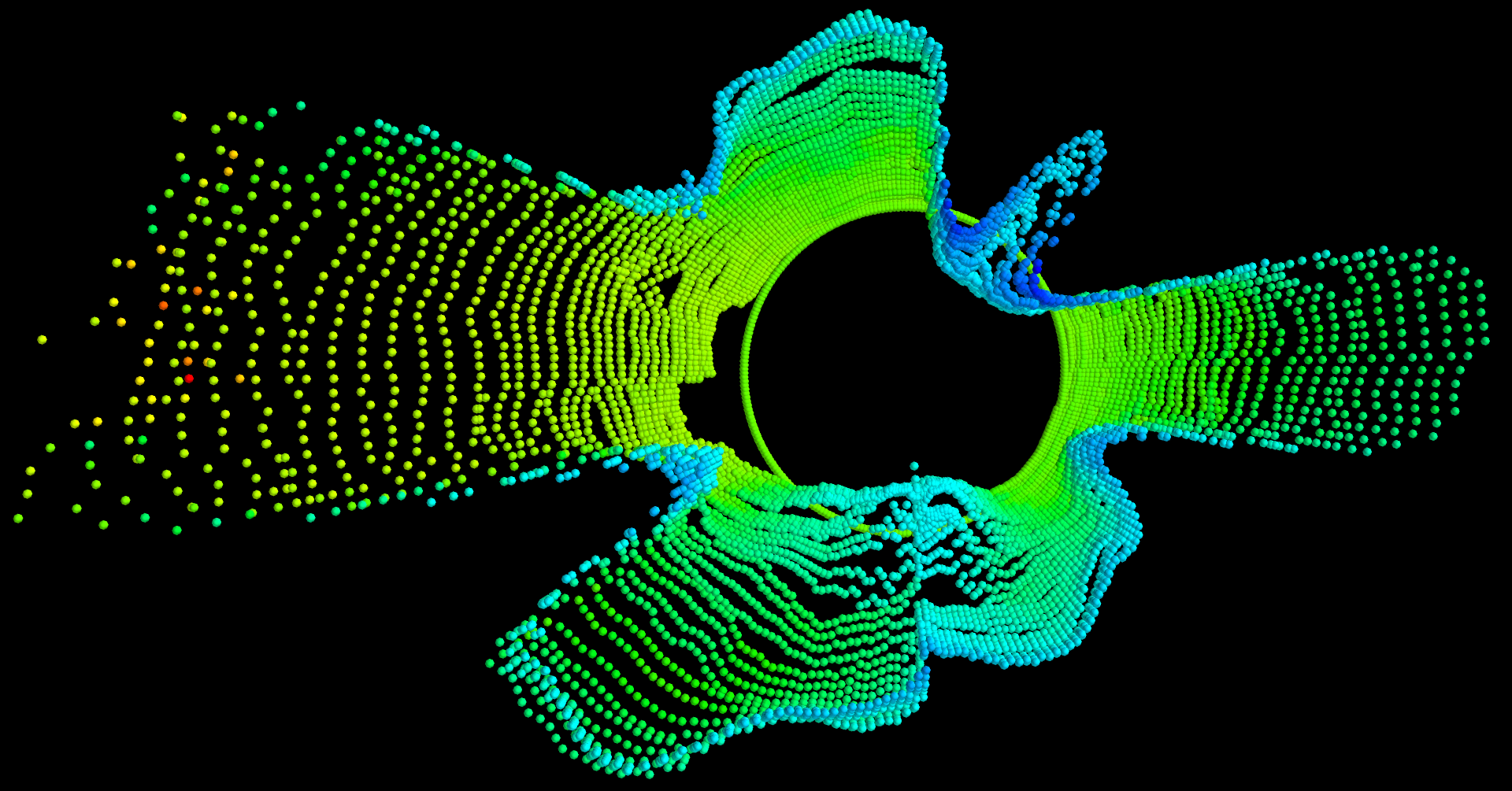}  
  \caption{standard LiDAR autoencoder backbones}
  \label{fig:f}
\end{subfigure}

\caption{Comparison of our Autoencoder module output \textit{w.r.t} a standard LiDAR autoencoder backbones. Column 1 represents original KITTI scan. Column 2 represents the reconstruction using our segmentation-attention assisted autoencoder which is coupled with contrastive learning. Column 3 represents the reconstruction using a standard LiDAR autoencoder backbones. Our autoencoder is able to reconstruct precise static and dynamic structures.}
\label{fig:kittiimages2}
\end{figure*}

We provide visualization and comparsion of the reconstruction results of $AE_{mask}$ against the best baselines in Figure \ref{kittiimages} and \ref{fig:kittiimages2}.
We provide a video demo of SLACK attacked KITTI sequence and its comparison with the original sequence. The video provided in the Supplementary material - SlackAttackedLiDAR-vs-originalLidarSequence.mp4. It is challenging to distinguish the attacked sequence from the original sequence in the video demo.

\section{Related Work}
\subsection{Adversarial Attacks on LiDAR data}
Several attacks have been proposed in the literature to destabilize navigation of LiDAR based AVs. Spoofing attacks use electromagnetic and laser interference to deceive the victim LiDAR sensor and either remove objects \cite{satoposter} or create fake objects \cite{wang2023adversarial}( object detection \cite{cho2023adopt}, segmentation \cite{johnson10lidar}) to attack the navigation pipeline. These use additional hardware mounted on an adversary vehicle or on roadside and attack the LiDAR during data collection. Saturation attacks \cite{jakobsen2023analysis} expose the target sensor to a signal beyond the lower or upper bound capacity making the sensor unable to work properly. Several adversarial attacks affect the latency of operation of an AV \cite{liu2023slowlidar}. Various white box attacks after gaining access to the AV module modify the range, timestamp and several other parameters of the datagram as well as add or drop datagrams \cite{hallyburton2023securing}. Several other attacks include forward replay, object removal attacks, Frustum Translation attacks. 

\subsection{Generative Modelling for Point Clouds}
\label{rel:gen}
Recent work on generative modelling for LiDAR, and in general for point clouds can be divided into 2 parts - \textbf{(1)} 3D shape point clouds: Several generative models for 3D shape based point clouds generation \cite{groueix2018papier,achlioptas2017representation} perform well on 3D shapes, but give sub-optimal results when adapted for multiple PiJ on LiDAR point clouds, due to the complexity of LiDAR data. The problem of incomplete 3D scan completion has been studied  by Wu et. al \cite{wu2020multimodal} and Chen et. al.\cite{chen2019unpaired}. Wu et. al. \cite{wu2020multimodal} generate multi-modal shapes for a given input but do not perform well when adapted for LiDAR scans \cite{kumar2021dslr}.
\textbf{(2)} LiDAR point clouds - Caccia et. al. \cite{caccia2018deep} generate conditional and unconditional generative models for LiDAR using polar range image representations. Zyrianov et. al. \cite{zyrianov2022learning} introduce LiDARGen that generates high quality samples using a score-matching based energy based model. It has a major disadvantage for SLAM - a sampling rate of 20s/scan. Xiong et. al. \cite{xiong2023learning} learn a discrete representation of a LiDAR. For manipulating the LiDAR objects, the authors rely on manual explicit intervention which is a drawback for end-to-end PiJ. Zhang et. al. \cite{zhang2023nerf} introduced 
 a NeRF based representation of scenes generate LiDAR scans. They requires multi-view images per LiDAR scan which is not available in our settings. The problem of dynamism manipulation has also been studied for LiDAR point clouds \cite{kumar2021dslr} and also for images \cite{bescos2019empty}, to some extent.
We show that existing feasible generative LiDAR modelling techniques when adapted for PiJ generate sub-optimal LiDAR quality on the LiDAR Quality Index (LQI) scale.

\subsection{Methodology}
Details on the segmentation-aware attention-based autoencoder backbone:  We employ a binary segmentation mask generated from LiDAR data, distinguishing between stationary and non-stationary points. This mask serves as a prior for reconstructing LiDAR data, particularly focusing on dynamic objects. The binary mask ensures focused attention on dynamic point features, facilitated by the segmentation encoder $H_{seg}$. Channel-level attention is then applied to the hidden layer features of $H_{phi}$ (the encoder of $AE_{mask}$) through adaptive average pooling over these features. 

Further to ensure that the parameters of $H_{seg}$ are tuned properly to induce proper attention to $AE$, we model another variant which trains $H_{seg}$ to reconstruct the segmentation mask using Dice coefficient loss.
We consider this loss instead of Binary Cross Entropy because dice coefficient loss is shown to work better at class imbalanced segmentation, which is the case here. Dynamic points are quite less in number compared to static points leading to imbalance between the classes. Finally the segmentation encoder ($H_{seg}$) network receives two signals during backpropagation, one each from the LiDAR autoencoder based $MSE$ loss and the segmentation encoder-decoder network using Dice coefficient loss.

Process of choosing multiple negatives: The LiDAR scans in a sequence are arranged in a contiguous fashion as they would appear while driving. We choose multiple negatives for the anchor as follows - we select the corresponding dynamic scan for the anchor scan as well as a couple of scans ahead and behind it in the driving sequence. 

The adversarial module takes heterogeneous pairs ($d_{j}, s_{j}$) as input and produces an adversarial output that assumes that the given input is a homogeneous pair. Among the pair, the segmentation mask (prior) that is fed to the adversarial module can be \textit{manipulated} as follows - The segmentation mask has the same size as the input range image. Portions of the mask are corrupted along the height of segmentation range image. This allows one to control the amount of dynamism and number of the dynamic locations injected into the static LiDAR scan. It discourage the mapping of the static latent representation to the same dynamic point on the dynamic manifold and allows for variety in \textbf{PiJ}.

\section{More SLAM Attack Results}

\begin{figure*}
\begin{subfigure}{.5\textwidth}
  \centering
  \includegraphics[width=\linewidth, height=\linewidth]{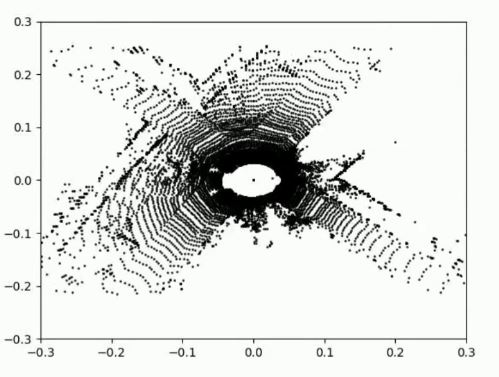}  
  \caption{Original Scan}
  \label{fig:sub-first}
\end{subfigure}
\begin{subfigure}{.5\textwidth}
  \centering
  \includegraphics[width=\linewidth, height=\linewidth]{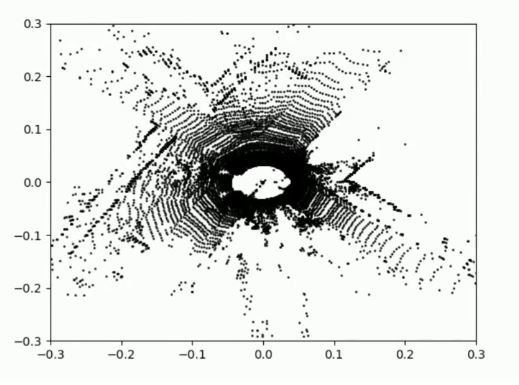}  
  \caption{Attacked Scan}
  \label{fig:sub-first}
\end{subfigure}
\caption{Visual demonstration of an original scan (left) and an attacked scan (right). It is very difficult to distinguish the attacked scan from the original scan. This is the objective of $AE_{seg}$ and SLACK. For a video demo, please refer to the video in the Supplementary.}
\label{fig:kitti-attack}
\end{figure*}

We provide a high-resolution image of Figure 6 in the main paper here (Figure \ref{kittiimages}). The figure demonstrates the efficacy of $AE_{seg}$ in maintaining the LiDAR quality. This in turn helps SLACK to maintain a desirable LiDAR Quality Index (lower is better) while also injecting LiDAR with \textit{PiJ}.

We demonstrate the effect of SLACK on the two CARLA-64 sequences in Figure \ref{fig:slam-carla-attack}. We observe that after the \textit{PiJ} attack with less than 1\% points, the generated map quality deteriorates significantly.  
The attacked map for sequence 1 does not accurately map the turns. The attacked map for sequence 2 fails to map some of the regions that are accurately mapped by the un-attacked LiDAR sequence. The trajectory generated after the attack has medium to high deviation from the ground truth trajectory (black color). The attacked trajectory for sequence 2 takes an unwanted turn and deviates significantly from the ground truth trajectory. 

It in interesting to note that these degradations in the map and trajectory estimates occur with less than 1\% \textit{PiJ}. 

We also demonstrate a comparison between the original and attacked LiDAR for the KITTI dataset in Figure \ref{fig:kitti-attack}. It is difficult to distinguish the attacked LiDAR scan from the original scan. For a video demo, please refer to the video in the Supplementary material.

\section{Baselines for $AE_{mask}$ and SLACK}

It has been established in literature that learning based methods significantly outperform geometric and other non learning based strategies\cite{kendall2015posenet,krull2015learning,bescos2019empty}. Moreover, vision based deep models work well using the latent space instead of image space, due to the fact that latent space aggregate only the dominant features of an image and are low-dimensional\cite{kumar2021dslr}.  Existing work on LiDAR generative models \cite{caccia2018deep},\cite{kumar2021dslr} also show that \cite{bescos2019empty,wu2020multimodal,groueix2018papier} do not perform well for LiDAR based generative modelling and hence have been omitted.

We use LiDAR based Autoencoder backbones that satisfy the following requirements necessary in a SLAM pipeline \\ 
\indent $\bullet$ A  high generation/sampling rate so that they are integratable in a SLAM pipeline.\\
\indent $\bullet$ Do not require datasets from modalities other than LiDAR point cloud.\\
\indent $\bullet$
Do not require manual intervention for \textit{PiJ} attacks.

We use \atlas (Groueix et al.\cite{groueix2018papier}), \Ach{},  (Achlioptas et. al.\cite{achlioptas2017representation}), \CacciaAE, \CacciaVAE, \CacciaGAN (Caccia et. al. \cite{caccia2018deep}), \DSLR (Kumar et. al. \cite{kumar2021dslr}) as baselines based on the above criteria.

Several recent works for generative modelling work very well and generate novel and realistic samples. In order to achieve this, they use data in additional modalities, complex methods (e.g. diffusion models), etc, which generate impressive results but break certain requirements which are necessary for end-to-end SLAM pipelines.
Existing works on LiDAR generative modeling \cite{caccia2018deep},\cite{kumar2021dslr} show that Bescos et al.\cite{bescos2019empty}, Wu et. al.\cite{wu2020multimodal} and Groueix et. al. \cite{groueix2018papier} do not perform well for LiDAR-based generative modeling and we do not compare against them. We are also unable to compare our work with Xiong et al. \cite{xiong2023learning}, Zyrianov et al \cite{zyrianov2022learning}, Zhang et. al. \cite{zhang2023nerf} due to the reasons explained below.

LiDARGen (Zyrianov et. al. \cite{zyrianov2022learning}) using diffusion models to develop an unconditional generative method for realistic LiDAR generation. Their sampling rate is extremely low to be able to be used in LiDAR SLAM pipelines.

UltraLiDAR (Xiong et. al.\cite{xiong2023learning}) learn a compact discrete representation for LiDAR which generates excellent novel samples. However, they require manual intervention for controllable manipulation which is a hindrance for an end-to-end pipeline for \textit{PiJ} attacks on LiDAR. \\

NeRF-LiDAR (Zhang et. al. \cite{zhang2023nerf}) use Neural Radiance Fields for simulation of novel real world LiDAR scans. However, they require multi-view images of a scene apart from LiDAR scans which are not available in our settings.

\section{Abation Studies}





\begin{table}[t]
\centering
\begin{tabular}{c|cccc}
\hline
Model & \multicolumn{2}{c|}{LQI} & \multicolumn{2}{c}{\makecell[1]{DSR}} \\
\hline
 & \multicolumn{1}{c|}{$AE_{mask}$} & \multicolumn{1}{c|}{\makecell[1]{Best\\ Baseline\\Backbone}}& \multicolumn{1}{c|}{$AE_{mask}$} & \multicolumn{1}{c}{\makecell[1]{Best\\w Baseline\\Backbone}}  \\
 \hline
 & \multicolumn{4}{c}{KITTI-64} \\
 \hline
\SAT{} & \multicolumn{1}{c|}{\textbf{1.97}} & \multicolumn{1}{c|}{3.24} & \multicolumn{1}{c|}{\textbf{0.48}} & \multicolumn{1}{c}{0.45} \\
\hline
 & \multicolumn{4}{c}{CARLA-64} \\
 \hline
\SAT{} & \multicolumn{1}{c|}{\textbf{3.73}} & \multicolumn{1}{c|}{4.48} & \multicolumn{1}{c|}{\textbf{0.51}} & \multicolumn{1}{c}{0.48} \\
\hline
\end{tabular}
\caption{Ablation studies to quantify the affect of segmentation aware LiDAR backbone - $AE_{mask}$ on \SAT{}. For LQI - lower is better. For DSR - higher is better.}

\label{tab:ablation_slack}
\end{table}

For $AE_{mask}$, we use 2 modules - segmentation-aware attention ($AE_{seg}$) and contrastive learning (Triplet and N-Pair Loss) using hard negatives. We perform Ablation studies to show the effect of these two modules in Table 
 \ref{tab:ablation_aeseg}. In the table, Caccia-AE refers to the best baseline backbone on top of which our segmentation-attention and contrastive module are augmented. We use the Caccia-AE backbone to augment our modifications because this backbone is straight forward to use and integrate our modules and works well with an impressive sampling/inference rate. Our modules can also be easily  integrated to other backbones.

We also show that for $AE_{mask}$, providing segmentation attention using the segmentation encoder to the LiDAR encoder (Figure 5a in main paper) is enough to encode the segmentation prior information into $AE_{mask}$. The segmentation encoder does not need a decoder to reconstruct the segmentation mask. On the contrary using an encoder-decoder segmentation network to reconstruct the segmentation information reduces the reconstruction quality. The ablation results in Table \ref{tab:ablation_aeseg} verify this claim.

We further show the benefit of using the $AE_{mask}$ module for SLACK over the best baseline backbone autoencoder in Table \ref{tab:ablation_slack}. We observe that segmentation-aware attention and contrastive learning using $AE_{mask}$ helps the network to maintain accurate LiDAR quality while performing \textit{PiJ} (Table \ref{tab:ablation_aeseg} and Figure \ref{fig:kitti-attack}). 

We also perform an ablation that demonstrates the importance of using the pretext task based discrimintor as opposed to using a standard discriminator for \textit{DiJ} in Table \ref{tab:ablation-disc}. 

\begin{table}[htbp]
\centering
\begin{tabular}{|c|c|c|}
\hline
 \textbf{Dataset} & \textbf{Vanilla Discriminator}& SLACK \\ \hline
 \multicolumn{3}{|c|}{{LQI/DSR}}         \\ \hline
 KITTI & 3.84/0.43  & \textbf{1.97/0.48}\\ \hline
 CARLA-64 & 5.47/0.50  & \textbf{3.73/0.51}\\ \hline
\end{tabular}
\caption{Ablation study to show the comparison between attack using a vanilla discriminator and SLACK.}
\label{tab:ablation-disc}
\end{table}

\section{Training and Experimental Setup}

We provide the details of the training setup for $AE_{mask}$ and SLACK. We train $AE_{mask}$ for 700 epochs. We use an Adam optimizer with an initial leanning rate of 0.001. The pretext task discriminator is trained for 20 epochs using the Adam optimizer with an initial learning rate of 0.0006 and a weight decay of 0.00001.
The adversarial module is trained  for 200 epochs using the Adam optimizer with an initial learning rate of 0.001 and a decay of 0.00001. 
For SLACK-MMD we initialize the network with pre-trained weights of $AE_{mask}$ and the pretext task discriminator, $PD$. The model is trained with KITTI data for Domain Adaptation till 200 epochs using Adam optimizer with an initial learning rate of 0.001 and a weight decay of 0.00001. 

All models are trained using an NVIDIA RTX-3090Ti GPU with 24 GB of GPU memory.  
All SLAM experiments are run on an Intel Core i7 CPU with 16 GB RAM. We use the ROS Noetic Distro for setting up Google cartographer for experiments with SLAM.

We will open source the codebase for $AE_{mask}$ and SLACK.

\section{Metrics}
\subsection{Evaluation of $AE{seg}$}
We describe the metrics for evaluating $AE_{mask}$.\\

\textbf{Earth Mover's Distance} - Given 2 LiDAR point clouds, $L_1$ and $L_2$, Earth Mover's Distance (EMD) calculates the least amount of work required to be done in order to transform $L_1$ to $L_2$ or vice-versa. Earth Mover's distance requires $L_1$ and $L_2$ to have the same number of points. EMD also requires a bijective mapping $\psi$ between the  point clouds. EMD is calculated as 
  \begin{equation}
    \begin{split}
        =\min _{\psi: L_{1} \longrightarrow L_{2}} \sum_{x \in L_{1}}\|x-\psi(x)\|_{2}
            \end{split}
        \end{equation}

\textbf{Chamfer's Distance} - Given 2 LiDAR point clouds $L_1$ and $L_2$, Chamfer's Distance between them is defined as 

\begin{equation}
            \begin{split}
          =\sum_{x \in L_{1}} \min _{y \in L_{2}}\|x-y\|_{2}^{2}+\sum_{x \in L_{2}} \min _{x \in L_{1}}\|x-y\|_{2}^{2}
          \end{split}
        \end{equation}

For every point $x \in L_1$, we find the Euclidean distance to the closest point in $L_2$ and vice-versa. The sum total of these distances is Chamfer's Distance between $L_1$ and $L_2$. Unlike EMD, Chamfer's Distance allows $L_1$ and $L_2$ to have different number of points. It does not necessitate the existence of a bijective mapping between $L_1$ and $L_2$. This allows a single point in $L_1$ to map
to multiple points in $L_2$, and vice versa.

\subsection{Evaluating of SLACK}
For evaluating SLACK we use 2 metrics - LiDAR Quality Index (LQI) and Dynamic Segmentation Ratio (DSR). DSR has been explained in the main paper. We give more details about LQI here.\\

\textbf{LQI} - LiDAR Quality Index or LQI, is used to asses the quality
of a given LiDAR scan. The prerequisite for using this
metric includes training a model with the domain scans
against which reconstructed test scans have to be evaluated. To evaluate LQI for a given LiDAR scan, it is passed through the trained model. The model regresses the amount of noise in the input. LQI is based on 
 work done on No-Reference Image Quality Assessment where the visual quality of an image is evaluated without any reference image or knowledge of the distortions present in the image \cite{kang2014convolutional}.

 The validation and usefulness of this metric is assessed by evaluating it against the original and reconstructed LiDAR scan of the domain. The
regressed noise (LQI) is less for the original LiDAR as compared to generated/reconstructed ones.

To evaluate LQI for a set of LiDAR
scans generated using a model M, we train another
model L using the original LiDAR scans - the original LiDAR scans are injected with noise
of varying levels. The injected noise is univariate Gaussian with mean 0 and variance $\sigma$.
The model L
is trained to be able to accurately regress the variance of the noise
present in the input LiDAR scan \cite{kumar2021dslr}.

\subsection{Metrics for evaluating SLAM}

To evalaute SLAM navigation  we use 2 metrics: Absolute Trajectory Error(ATE) and Relative Pose Error(RPE) 

\textbf{Absolute Trajectory Error} - ATE is used to measure the overall global consistency between two trajectories. It
measures the difference between the translation components of the two trajectories using a 2 step process - aligning the coordinate frames for both trajectories 
 and then comparing the absolute distances
between the translational components of the individual poses. For more details on the procedure for calculating ATE, please refer to Sturm et. al. \cite{sturm2012benchmark}.

\textbf{Relative Pose Error}: RPE measures the deviation of the poses between the estimated and the ground truth trajectory locally - it measures the discrepancy between the estimated
and ground truth trajectories over a specified time interval. It is robust to the accumulated global error. RPE takes the
translational and rotational components of the trajectory into account and calculates the pose error separately for both. For a detailed description of the error and the calculation procedure, please refer to the Sturm et. al. \cite{sturm2012benchmark}.

\subsection{Datasets}
\label{dataset}
 \textit{CARLA-64} - CARLA-64 is a simulated 64-beam LiDAR dataset collected using the setting of a Velodyne VLP-64 LiDAR sensor on CARLA simulator \cite{dosovitskiy2017carla}. CARLA-64 consists of $\sim$ 32,000 corresponding static-dynamic LiDAR scan pairs along with the static-dynamic point segmentation mask information. We use the latest version of the dataset and the mentioned train/test protocol \cite{dslr-git} for testing \textit{PiJ} using SLACK. We use the two available SLAM sequences for testing the effect of PiJ attacks on SLAM.
    
\textit{KITTI-64} - To validate our results on a real-world dataset where corresponding scans are not available we consider KITTI 64-beam LiDAR dataset \cite{geiger2012we}. KITTI has segmentation masks available. We demonstrate PiJ attacks using SLACK on the KITTI Odometry sequences 00-10. We also use the Odometry sequences for testing the affect of SLACK on SLAM.

\textit{ARD-16} - We also test PiJ using SLACK on a real-world 16-beam industrial dataset, collected using a VLP-16 Puck LiDAR sensor on an industrial robot \cite{dslr-git}. We use this dataset to study the behavior of our model on sparse datasets. ARD-16 has correspondence pairs available. It does not have segmentation labels available.



\section{Ablation Studies}
\label{sec:ablation}
We quantify the importance of segmentation-based Attention and contrastive learning for $AE_{mask}$ using ablation studies. 
We also demonstrate the benefit of using $AE_{mask}$ backbone with SLACK as opposed to using a standard LiDAR autoencoder backbone.

\begin{table}[t]
\centering
\begin{tabular}{c|cc}
\hline
Model & \multicolumn{2}{c}{CARLA-64} &  \hline
 & \multicolumn{1}{c}{Chamfer} & EMD & 
\Xhline{3\arrayrulewidth}
Caccia-AE & \multicolumn{1}{c|}{1.52} & 164 \\

\hline
Caccia-AE+TripletLoss & \multicolumn{1}{c|}{1.30} & 164 \\ \hline
Caccia-AE+N-pairLoss & \multicolumn{1}{c|}{0.98} & 139 \\ \hline

$AE_{mask}$ & \multicolumn{1}{c|}{1.00} & 153 \\ \hline
$AE_{mask}$+TripletLoss & \multicolumn{1}{c|}{1.20} & 157\\ \hline

\Xhline{2\arrayrulewidth}
( $AE_{mask}$+N-pair Loss) - \textbf{Ours} & \multicolumn{1}{c|}{\textbf{0.93}} & \textbf{126}\\ \hline

\end{tabular}
\caption{Ablation studies to study the effect of segmentation-aware attention and constrastive losses for $AE_{seg}$.}
\label{tab:ablation_aeseg}
\end{table}

\begin{table}[t]
\centering
\begin{tabular}{c|cccc}
\hline
Model & \multicolumn{2}{c|}{LQI} & \multicolumn{2}{c}{\makecell[1]{DSR}} \\
\hline
 & \multicolumn{1}{c|}{$AE_{mask}$} & \multicolumn{1}{c|}{\makecell[1]{Best\\ Baseline\\Backbone}}& \multicolumn{1}{c|}{$AE_{mask}$} & \multicolumn{1}{c}{\makecell[1]{Best\\w Baseline\\Backbone}}  \\
 \hline
 & \multicolumn{4}{c}{KITTI-64} \\
 \hline
\SAT{} & \multicolumn{1}{c|}{\textbf{1.97}} & \multicolumn{1}{c|}{3.24} & \multicolumn{1}{c|}{\textbf{0.48}} & \multicolumn{1}{c}{0.45} \\
\hline
 & \multicolumn{4}{c}{CARLA-64} \\
 \hline
\SAT{} & \multicolumn{1}{c|}{\textbf{3.73}} & \multicolumn{1}{c|}{4.48} & \multicolumn{1}{c|}{\textbf{0.51}} & \multicolumn{1}{c}{0.48} \\
\hline
\end{tabular}
\caption{Ablation studies to quantify the affect of segmentation aware LiDAR backbone - $AE_{mask}$ on \SAT{}. For LQI - lower is better. For DSR - higher is better.}

\label{tab:ablation_slack}
\end{table}

\subsubsection{$AE_{mask}$}
For $AE_{mask}$, we use 2 modules - segmentation-aware attention ($AE_{seg}$) and contrastive learning (Triplet and N-Pair Loss) using hard negatives. We perform Ablation studies to show the effect of these two modules in Table 
 \ref{tab:ablation_aeseg}. In the table, Caccia-AE refers to the best baseline backbone on top of which our segmentation-attention and contrastive module are augmented. We use the Caccia-AE backbone to augment our modifications because this backbone is straight forward to use and integrate our modules and works well with an impressive sampling/inference rate. Our modules can also be easily  integrated to other backbones.

We also show that for $AE_{mask}$, providing segmentation attention using the segmentation encoder to the LiDAR encoder (Figure 5a in main paper) is enough to encode the segmentation prior information into $AE_{mask}$. The segmentation encoder does not need a decoder to reconstruct the segmentation mask. On the contrary using an encoder-decoder segmentation network to reconstruct the segmentation information reduces the reconstruction quality. The ablation results in Table \ref{tab:ablation_aeseg} verify this claim.

We further show the benefit of using the $AE_{mask}$ module for SLACK against the best backbone autoencoder baseline, in Table \ref{tab:ablation_slack}. We observe that segmentation-aware attention and contrastive learning using $AE_{mask}$ helps the network to maintain accurate LiDAR quality while performing \textit{PiJ} (Table \ref{tab:ablation_aeseg} and Figure \ref{fig:kitti-attack}). 

We also perform an ablation that demonstrates the importance of using the pretext task based discrimintor for SLACK as opposed to using a standard discriminator for \textit{DiJ} in Table \ref{tab:ablation-disc}. 

\begin{table}[htbp]
\centering
\begin{tabular}{|c|c|c|}
\hline
 \textbf{Dataset} & \textbf{Vanilla Discriminator}& SLACK \\ \hline
 \multicolumn{3}{|c|}{{LQI/DSR}}         \\ \hline
 KITTI & 3.84/0.43  & \textbf{1.97/0.48}\\ \hline
 CARLA-64 & 5.47/0.50  & \textbf{3.73/0.51}\\ \hline
\end{tabular}
\caption{Ablation study to show the comparison between attack using a vanilla discriminator and SLACK.}
\label{tab:ablation-disc}
\end{table}


\bibliographystyle{IEEEbib}
\bibliography{refs}